%% file: main.tex
\pdfoutput=1
\documentclass[10pt,twocolumn,letterpaper]{article}

\usepackage[pagenumbers]{cvpr}
\makeatletter\patchcmd{\@maketitle}{\vskip .375in}{\vskip 5pt}{}{\PackageError{arxiv}{title patch failed}{}}
\patchcmd{\@maketitle}{\vspace*{24pt}}{\vspace*{19pt}}{}{\PackageError{arxiv}{title patch failed}{}}
\patchcmd{\@maketitle}{\vspace*{12pt}}{\vspace*{6pt}}{}{\PackageError{arxiv}{title patch failed}{}}
\def\arxiv@suppaux@input{\@input{supp.aux}}\makeatother

\input{preamble}

\definecolor{cvprblue}{rgb}{0.21,0.49,0.74}
\usepackage[pagebackref,breaklinks,colorlinks,allcolors=cvprblue]{hyperref}
\hypersetup{pdftitle={BEND: Spectral Bregman Losses for Multi-Image Unlearning in Diffusion Models}, pdfauthor={Mohamed-Hicham Leghettas, Markus P\"uschel}}

\title{BEND: Spectral Bregman Losses for Multi-Image Unlearning in Diffusion Models}

\author{Mohamed-Hicham Leghettas \qquad Markus P\"uschel\\
Department of Computer Science, ETH Zurich\\
{\tt\small \{mleghettas, pueschel\}@inf.ethz.ch}}

\begin{document}

\maketitle
\addtocontents{toc}{\protect\setcounter{tocdepth}{-10}}

\begin{abstract}
Diffusion models memorize training images and reproduce them almost verbatim when prompted, creating copyright and privacy liabilities. Data unlearning erases the influence of specific training samples from a trained model without retraining, and is two-sided: it must forget the targets while preserving generation quality on everything else. Prior methods use the squared-$L^2$ diffusion loss, at most with a hand-set frequency mask; however, it is known that every Bregman divergence is a mathematically valid choice and the squared-$L^2$ is only the simplest example. BEND is the first to \emph{search} this family: it fits a spectral Bregman divergence to the unlearning task through a bilevel loop, so the loss itself is shaped to forget thoroughly while protecting quality. It also redirects a memorized image to the mean of its retain images, where ReTrack, the state of the art, redirects it to a single near-copy. On Stable Diffusion, a \emph{single} BEND model erases all $48$ verifiably memorized benchmark images at once. It forgets at least as much as per-target, sequential, and jointly trained ReTrack (SSCD $0.084$ versus $0.094$--$0.120$), at the quality of the model before unlearning (CLIP-IQA $0.81$ versus $0.64$--$0.69$). BEND also outperforms the state of the art on CelebA-HQ in all three measures of the benchmark, and matches or exceeds it on MNIST and CIFAR-10.
\end{abstract}

\section{Introduction}
\label{sec:intro}

\looseness=-1 Diffusion models for image generation are trained by scraping enormous image datasets. They not only learn the underlying distribution, they also memorize individual training images and reproduce them almost verbatim when prompted \citep{carlini2023extracting, somepalli2023diffusion}. This creates a liability when such an image is copyrighted or shows a recognizable private face, since data-protection regulation increasingly grants the right to have such content removed: for example, regulators have ordered the deletion of residents' faces from a scraped database (Clearview AI, under the GDPR \citep{cnil2022clearview}) and even of models trained on users' photos (Everalbum, by the U.S.\ FTC \citep{ftc2021everalbum}). Retraining from scratch to honor each removal request is prohibitively expensive. \emph{Data unlearning} instead edits an already-trained model to erase the influence of specific samples with limited additional compute. The task is inherently two-sided: the model must forget the target samples while leaving its generation quality on everything else intact.

\looseness=-1 Data-unlearning methods use the squared-$L^2$ noise-prediction loss, and differ mainly in what they feed it: NegGrad \citep{golatkar2020neggrad} ascends it on the image to be forgotten, EraseDiff \citep{wu2024erasediff} aims it at surrogate noise targets, SISS \citep{alberti2025siss} mixes descent and ascent by importance sampling, and ReTrack \citep{shi2026retrack}, the current state of the art, changes only which noisy samples are drawn. The loss itself has at most been masked by hand \citep{park2025frequency}; it has never been searched.

Yet the loss is a free design choice: a diffusion loss is valid, in the precise mathematical sense that minimizing it also minimizes the intractable training objective it substitutes for, \emph{if and only if} it is a Bregman divergence, generated by a strictly convex potential $\Phi$ \citep{banerjee2005bregman, lipman2024fmguide}. The squared-$L^2$ loss is a particular case corresponding to the potential $\Phi(\vu) = \tfrac{1}{2}\|\vu\|_2^2$; every other strictly convex $\Phi$ yields a mathematically equally valid loss with a different geometry.

\begin{figure*}[t]
	\centering
	\includegraphics[width=\columnwidth]{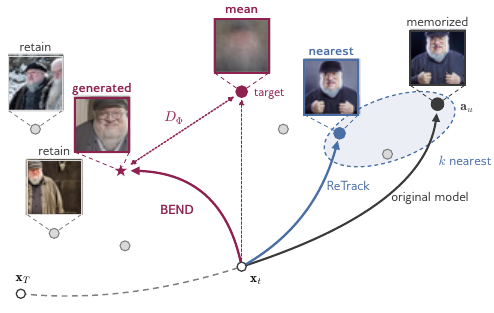}\hfill
	\includegraphics[width=\columnwidth]{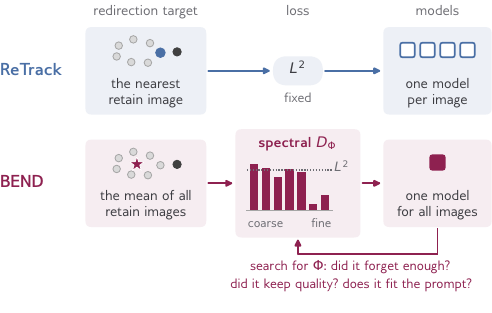}
	\caption{BEND at a glance. \emph{Left}: where unlearning redirects a noisy sample $\vx_t$ of a memorized prompt (images of one Stable Diffusion target): ReTrack to the nearest retain image, here a near-copy of the memorized image $\va_u$; BEND to the mean of all retain images, but the model ends at a new, sharp image ($\star$), which $D_\Phi$ counts as close to the mean although the squared-$L^2$ does not. \emph{Right}: the two methods side by side; the bars give the weight of each spatial scale in $D_\Phi$. Prior methods fix the unlearning loss; BEND \emph{searches} for it.}
	\label{fig:concept}
\end{figure*}

\looseness=-1 In this paper we present the data-unlearning method BEND, sketched in Fig.~\ref{fig:concept}. Instead of using the standard squared-$L^2$ loss, BEND searches the Bregman family for a divergence: it parameterizes $\Phi$ by how much each spatial scale and each noise level of the prediction error counts and tailors the divergence $\Phi$ generates to the unlearning task through a bilevel loop. An inner loop unlearns the model under the current loss; an outer loop adjusts the loss so the resulting model forgets and retains well. The outer loop scores what matters in the end, the images the unlearned model generates: how similar they are to the memorized image, how good they look, and whether they still fit the prompt. None of these measurements is differentiable, so the outer loop works without gradients, at the cost of a few dozen short unlearning runs.

\looseness=-1 BEND builds on ReTrack, which forgets a memorized image by redirecting its denoising toward nearby retain images. On Stable Diffusion we found that this redirection collapses onto a single image, and that this image is on average twice as similar to the memorized image as the retain images are. BEND therefore redirects to the mean of the retain images instead. With this target and the searched loss, multiple images can be forgotten jointly by one model, where prior work trains one model per image.

We offer the following contributions:
\begin{itemize}
\item We identify the per-sample loss, determined by the choice of Bregman potential $\Phi$, as a design choice for data unlearning in diffusion models that can be searched rather than set by hand.
\item We leverage this observation with BEND, which searches a spectral family of Bregman losses for one suited to a given forgetting task with a bilevel loop whose outer objective is measured on the images the unlearned model generates.
\item We show that the redirection target of ReTrack degenerates on Stable Diffusion to a single near-copy of the memorized image, and replace it by the mean of the retain images, the minimizer of every Bregman loss over them.
\item We evaluate BEND on all $48$ verifiably memorized images of the Stable Diffusion benchmark. One BEND model forgets more than per-target and sequential ReTrack, and as much as jointly trained ReTrack, at substantially higher quality. BEND further outperforms the state of the art on CelebA-HQ and matches or exceeds it on MNIST and CIFAR-10.
\end{itemize}

\section{Data Unlearning and Bregman Losses}
\label{sec:bregman}

We review diffusion training and the Bregman family of losses, then state the data-unlearning problem. We are the first to search Bregman losses for data unlearning.

\mypar{Diffusion training}
A diffusion model $\vepsilon_{\vtheta}$ is trained to predict the noise added to a clean sample $\va \in \sA$ from a dataset $\sA \subset \R^d$. The forward process supplies the training pairs: it gradually turns a clean sample $\va$ into noise. At timestep $t \in \{1, \dots, T\}$ (with $T = 1000$ in all our models) it scales $\va$ by $\gamma_t \in \R$ and adds standard Gaussian noise $\vepsilon \sim \Normal(\vzero, \mI_d)$ scaled by $\sigma_t > 0$, producing the noisy sample $\vx_t = \gamma_t \va + \sigma_t \vepsilon$. The schedule $(\gamma_t, \sigma_t)$ fixes how much signal and noise $\vx_t$ carries; equivalently, $\vx_t \sim q_t(\cdot \mid \va) = \Normal(\gamma_t \va,\, \sigma_t^2 \mI_d)$. Given only $\vx_t$ and $t$, the model predicts the added noise; the noise target consistent with a clean sample $\va$ is $\vepsilon_t(\va) = (\vx_t - \gamma_t \va)/\sigma_t$. Denoising diffusion probabilistic models (DDPMs) are trained with the squared-$L^2$ loss against this target,
\begin{equation}
\ell_t(\vtheta;\, \va) = \big\|\vepsilon_{\vtheta}(\vx_t, t) - \vepsilon_t(\va)\big\|_2^2
\label{eq:residual}
\end{equation}
where the dependence of $\vepsilon_t(\va)$ and $\ell_t$ on the realized $\vx_t$ and $t$ is left implicit. When $\vx_t$ is drawn from the forward process the noise target coincides with the added noise, $\vepsilon = \vepsilon_t(\va)$, so the standard DDPM objective \citep{ho2020ddpm} is simply the expectation of the squared-$L^2$ loss $\ell_t(\vtheta;\, \va)$.

\mypar{Bregman divergence}
Let $\Phi : \R^d \to \R$ be strictly convex and differentiable. The Bregman divergence \citep{bregman1967relaxation} generated by $\Phi$,
\begin{equation}
D_\Phi(\va, \vb) = \Phi(\va) - \Phi(\vb) - \langle \nabla \Phi(\vb),\, \va - \vb \rangle,
\label{eq:bregman}
\end{equation}
is the gap between $\Phi(\va)$ and the first-order Taylor expansion of $\Phi$ about $\vb$. Strict convexity makes this gap non-negative and zero only at $\va = \vb$. In general $D_\Phi$ is asymmetric, and the order of its arguments matters. The canonical example is the squared-$L^2$ loss, generated by $\Phi(\vu) = \tfrac{1}{2}\|\vu\|_2^2$ and giving $D_\Phi(\va, \vb) = \tfrac{1}{2}\|\va - \vb\|_2^2$; the Kullback--Leibler and Itakura--Saito divergences are other classical examples \citep{banerjee2005bregman}.

\mypar{The valid loss family}
Diffusion training is a special case of the flow-matching framework \citep{lipman2023flowmatching, lipman2024fmguide}. Flow-matching training is valid \emph{if and only if} the per-sample loss is a Bregman divergence, with the ``if'' shown by \citet[Theorem~4]{lipman2024fmguide} and the ``only if'', under mild regularity, by \citet[Theorems~3--4]{banerjee2005bregman}:
\begin{equation}
\ell_t(\vtheta;\, \va) = D_\Phi(\vepsilon_t(\va), \vepsilon_{\vtheta}(\vx_t, t))
\label{eq:bregman-loss}
\end{equation}
for some strictly convex $\Phi$; the argument is given in the supplementary material. \emph{Valid} here means the loss we can actually compute, one example at a time, has the same gradient as the true objective it substitutes for, so minimizing the cheap one minimizes the real one. The reason is a property that characterizes the family: when the target is random, the expected Bregman loss is minimized by the \emph{mean} of the target, for every $\Phi$. All valid losses thus share the same optimum, but not the path to it: the gradient of $D_\Phi$ is the error multiplied by the Hessian of $\Phi$, so $\Phi$ decides which parts of the error are corrected first. This matters for unlearning, which is a short fine-tuning run that stops long before any optimum is reached. The squared-$L^2$ loss is the trivial case $\Phi(\vu) = \tfrac{1}{2}\|\vu\|_2^2$. Other choices have at most been set by hand in this context \citep{park2025frequency}.

\mypar{Data unlearning}
Data unlearning removes the influence of specific training samples from an already-trained model. Given $\vepsilon_{\vtheta}$ pretrained on $\sA$, it seeks updated parameters such that the resulting model no longer carries the influence of the forget set $\sA_u \subset \sA$ (also called the \emph{unlearning set}) and behaves as though it had been trained only on the retain set $\sA_r = \sA \setminus \sA_u$. In prior work, each request concerns a single image, $\sA_u = \{\va_u\}$. Success is therefore two-sided: the model must erase $\sA_u$ while preserving generation quality on the retain distribution.

\section{Related Work}
\label{sec:sota}

\looseness=-1 We review the data-unlearning baselines benchmarked by \citet{shi2026retrack}, then ReTrack itself. We omit VDU \citep{panda2024vdu}, which needs several model checkpoints to estimate a posterior over the parameters, and concept- and class-erasure methods (ESD~\cite{gandikota2023esd}, Selective Amnesia~\cite{heng2023selective}, recent variants~\cite{bui2025fantastic, gao2025eraseanything}, and the diffusion experiments of SalUn~\cite{fan2024salun}), which target a prompt- or class-defined forget set rather than a training-sample list and are evaluated on disjoint benchmarks. Memorization can also be mitigated without unlearning, during training or at inference \citep{somepalli2023understanding, wen2024detecting, chen2024memorizationfree, ren2024unveiling}.

\mypar{Vanilla}
The natural baseline fine-tunes the pretrained model on $\sA_r$ alone with the original DDPM loss
\begin{equation}
J_{\mathrm{van}}(\vtheta) = \E_{\substack{t, \va_r \sim \sA_r \\ \vx_t \sim q_t(\cdot \mid \va_r)}}[\ell_t(\vtheta;\, \va_r)].
\label{eq:vanilla}
\end{equation}
Its optimum matches a model retrained from scratch on $\sA_r$, but its gradient concentrates on retain inputs, where the model already fits well; forgetting $\sA_u$ is far too slow in practice.

\mypar{NegGrad}
NegGrad \citep{golatkar2020neggrad}, applied to diffusion by \citet{alberti2025siss}, ascends the diffusion loss on the forget image, i.e., it descends $J_{\mathrm{NG}}(\vtheta)$, the vanilla objective~\eqref{eq:vanilla} over $\sA_u$ with its sign flipped. Forgetting is fast, but the objective is unbounded below and unconstrained on $\sA_r$; since forget and retain inputs share parameters, retain-set quality collapses as the model is pushed off $\sA_u$.

\mypar{EraseDiff}
EraseDiff \citep{wu2024erasediff}, adapted to data unlearning by \citet{alberti2025siss}, keeps the vanilla cost on $\sA_r$ but pushes the noise prediction on the forget image toward a uniform-noise target, with a per-step adaptive weight \citep{liu2022bome}. Aiming at a fixed target rather than ascending away from the true noise keeps forgetting bounded, avoiding NegGrad's collapse, but the surrogate target biases the optimum away from the retraining solution.

\mypar{SISS}
SISS \citep{alberti2025siss} draws each noisy sample from a mixture of the retain- and forget-image-conditional distributions and reweights it by importance sampling, so a single sample serves both branches of the loss. In expectation the objective equals $J_{\mathrm{van}}(\vtheta) + s\, J_{\mathrm{NG}}(\vtheta)$ for a weight $s$ \citep{shi2026retrack}: the shared sample halves the per-step cost \citep{alberti2025siss}, but the forget signal is still NegGrad's biased ascent term. ReTrack gives a better trade-off than all three \citep{shi2026retrack}, so our comparison concentrates on it.

\mypar{ReTrack}
\looseness=-1 ReTrack \citep{shi2026retrack} instead draws the noisy sample near the forget sample $\va_u$ but keeps retain neighbors $\va_r$ as the denoising targets, leaving the squared-$L^2$ loss form unchanged:
\begin{equation}
J_{\mathrm{unl}}(\vtheta) = \E_{\substack{t, \va_u \sim \sA_u \\ \vx_t \sim q_t(\cdot \mid \va_u)}}\Big[\sum_{\va_r \in \sA_r} w_t(\vx_t \mid \va_r)\, \ell_t(\vtheta;\, \va_r)\Big]
\label{eq:retrack-unl}
\end{equation}
with weights $w_t(\vx_t \mid \va_r) \propto q_t(\vx_t \mid \va_r)$, normalized over $\sA_r$. Sampling near $\va_u$ concentrates the gradient on exactly the inputs the model has memorized, the region vanilla leaves untouched, while the retain-neighbor targets teach it to denoise a corrupted $\va_u$ as if it had come from nearby real data, redirecting that sample's denoising trajectory back onto the data manifold. Because only the sampling changes and not the loss, the objective still shares vanilla's optimum \citep[Prop.~1]{shi2026retrack}: unlearning adds no bias of its own. To keep the inner sum tractable it is truncated to the $k$ nearest neighbors of $\va_u$, and a vanilla regularizer guards the retain set:
\begin{equation}
J_{\mathrm{ReTrack}}(\vtheta) = \lambda\, J_{\mathrm{unl}}(\vtheta;\, k) + (1 - \lambda)\, J_{\mathrm{van}}(\vtheta).
\label{eq:retrack}
\end{equation}
Unlike NegGrad, EraseDiff, and SISS, ReTrack forgets through this change of sampling rather than a biased term: no surrogate target or ascent term appears.

\mypar{Loss design}
\looseness=-1 \citet{park2025frequency} mask the forget loss of several methods with a hand-designed low-pass filter and a timestep window. Such a mask is a member of the family we search, with its weights set by hand; our hand-set ablation, which closes the two finest frequency bands at every noise level, is close to it. Weighting the diffusion loss by noise level is common in training \citep{choi2022p2, hang2023minsnr, karras2022edm, kingma2023understanding}, and our exponent $p$ is such a weighting; frequency-weighted losses are used in image synthesis \citep{jiang2021focal}. Targets built from generations of an anchor prompt, or from a tilted distribution, appear in concept erasure \citep{kumari2023ablating, george2026tilde} and in KL-constrained unlearning \citep{khalafi2026erasure}, and learning a loss by meta-optimization is established in other settings \citep{bechtle2021learnedloss}.

\section{The BEND Method}
\label{sec:method}

\looseness=-1 In Section~\ref{sec:bregman} we asserted that a diffusion loss is valid exactly if it is a Bregman divergence, and that every such divergence is generated by a potential function $\Phi$. Prior methods all fix $\Phi$ to the single choice that yields the squared-$L^2$ loss. BEND instead treats the potential as something to be \emph{found}: we search for the $\Phi$ whose divergence is best suited to the unlearning task, and then unlearn with it (Fig.~\ref{fig:concept}). The unlearning procedure itself starts from ReTrack, the current state of the art, with one correction to what its forget term redirects to.

\begin{figure}[t]
	\centering
	\includegraphics[width=\columnwidth]{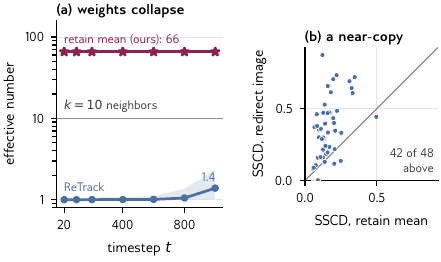}
	\caption{What ReTrack redirects to, on the $48$ memorized Stable Diffusion images. (a) Effective number of retain images behind the redirect image, by timestep (band: range over targets). (b) SSCD to the memorized image of the redirect image against the mean of the target's retain images; one point per target.}
	\label{fig:redirect}
\end{figure}

\mypar{What ReTrack redirects to}
\looseness=-1 The forget term~\eqref{eq:retrack-unl} regresses the prediction at a noisy forget sample onto the noise targets of several retain images at once. Since every Bregman loss is minimized by the mean of its targets (Section~\ref{sec:bregman}), this is the same as regressing onto one target, the noise target of the weighted average image $\sum_r w_t(\vx_t \mid \va_r)\, \va_r$. We call it the \emph{redirect image}: it is what the model is taught to produce in place of the memorized image. Intuitively, this goes wrong when the retain set contains a near-duplicate of the memorized image, which is likely, since memorization is driven largely by duplication: memorized images tend to occur many times in the training data, in identical or slightly varied copies \citep{carlini2023extracting, somepalli2023diffusion}. Deleting the memorized image leaves its near-duplicates in the retain set, and ReTrack, whose weights favor the retain image closest to the noisy sample, can redirect the model to one of them (Fig.~\ref{fig:concept}, left). We measured the redirect image on Stable Diffusion (Fig.~\ref{fig:redirect}), and it is not the blend of plausible alternatives one would expect. The weights are degenerate: the effective number of retain images that carry weight, $1/\sum_r w_r^2$, is $1.00$ up to $t = 400$, $1.05$ at $t = 800$, and reaches only $1.4$ at the last timestep. The reason is the dimension of the latent, $16{,}384$, which makes squared distances differ far too much between two images for both to keep weight (supplementary material); truncating the sum to $k$ neighbors is not the cause. The one image that survives is the retain image closest to the memorized one, and closest in the latent space means most similar in content. Its SSCD similarity to the memorized image is $0.36$ on average, against $0.17$ for the retain images as a whole, and it is the more similar of the two for $42$ of the $48$ targets. For $10$ targets it even exceeds the SSCD of $0.5$ above which \citet{somepalli2023diffusion} observe significant copying, although the benchmark's clustering counts it as a retain image.

\mypar{Redirecting to the retain images as a whole}
\looseness=-1 BEND drops the weights and redirects to the retain images as a whole. Let $\sS \subset \sA_r$ be the retain images that belong to a forget target; on Stable Diffusion these are the non-memorized generations for its prompt (Section~\ref{sec:setup}). The target of the forget term is the mean of their noise targets,
\begin{equation}
\bar\vepsilon_t = \frac{1}{|\sS|}\sum_{\va_r \in \sS} \vepsilon_t(\va_r) = \frac{\vx_t - \gamma_t\, \bar\va}{\sigma_t},
\qquad \bar\va = \frac{1}{|\sS|}\sum_{\va_r \in \sS} \va_r .
\label{eq:poolmean}
\end{equation}
For every Bregman loss this is equivalent to drawing a fresh retain image for each noisy sample, only with less variance. It is also what a denoiser that never memorized $\va_u$ predicts for the prompt at high noise. If an application demands a stricter forgetting level, the retain images can additionally be reweighted away from the memorized image by an exponential tilt; we evaluate this variant in the supplementary material.

\mypar{The spectral loss}
\looseness=-1 The mean image $\bar\va$ is blurry: it carries the composition that the retain images share, but none of their detail. Asking the model to match it at every scale, as the squared-$L^2$ loss does, forgets well but teaches the model to blur (Section~\ref{sec:results}). The choice of loss decides whether this happens: under a loss that compares mainly coarse scales, a sharp image with the composition of $\bar\va$ is close to it, although the squared-$L^2$ counts it as far (Fig.~\ref{fig:concept}, left). BEND's loss is a single searched Bregman divergence that replaces the squared-$L^2$ loss $\ell_t$ in the forget term of ReTrack's objective~\eqref{eq:retrack}:
\begin{equation}
\ell_t^{\mathrm{BEND}}(\vtheta) = D_{\Phi_\psi}\big(\bar\vepsilon_t,\, \vepsilon_{\vtheta}(\vx_t, t);\, t\big).
\label{eq:bend-loss}
\end{equation}
Its generator $\Phi_\psi$ weighs the error by spatial scale and by noise level. Writing $\hat\vu$ for the two-dimensional discrete Fourier transform of $\vu$ over the spatial axes,
\begin{equation}
\Phi_\psi(\vu;\, t) = \frac{1}{2} \sum_{k} w_\psi\big(t, b(k)\big)\, |\hat u_k|^2,
\label{eq:potential}
\end{equation}
where $b(k)$ is the octave band of frequency $k$ (seven bands, from the coarsest scales to the finest) and the weights are positive. Since $\Phi_\psi$ is quadratic, $D_{\Phi_\psi}$ is a squared-$L^2$ weighted in the Fourier basis; we search this spectral sub-family and leave non-quadratic potentials open. The parameters $\psi$ are one gate $g_b$ per band and an exponent $p$ that shifts weight between low and high noise levels,
\begin{equation}
\log w_\psi(t, b) = g_b - p\,\big(\ell_t - \bar\ell\,\big) + c(\psi),
\label{eq:gates}
\end{equation}
with $\ell_t = \log(\gamma_t^2/\sigma_t^2)$ the log signal-to-noise ratio and $\bar\ell$ its mean. The constant $c(\psi)$ is a normalization that keeps the overall weight of the forget term at that of the squared-$L^2$ loss, so $\psi$ changes the shape of the loss and not its balance against the retain term. Positivity of the weights makes $\Phi_\psi$ strictly convex, which is what makes $D_{\Phi_\psi}$ a valid loss for every $\psi$ (Section~\ref{sec:bregman}), keeping the search inside the valid family. The potential is deliberately small: with eight parameters it can be found from measurements that provide no gradient, and each parameter can be read afterwards. The retain term keeps the standard loss. Since the target and this substitution are the only changes, with the near-forget sampling untouched, they alone define the inner objective $\mathcal{L}_{\mathrm{inner}}(\vtheta;\, \psi)$, whose forget term recovers the squared-$L^2$ loss exactly if $\psi = 0$.

\mypar{Finding a good $\Phi$}
\looseness=-1 A potential $\Phi$ cannot be judged on its own; what makes it good is the diffusion model that results from unlearning with it. We therefore judge a candidate $\Phi$ (one setting of the parameters $\psi$) by running unlearning with it and scoring the model it produces. Running unlearning is the \emph{inner loop}: it fine-tunes the diffusion weights $\vtheta$ under the divergence $\Phi$ generates, yielding an unlearned model $\vtheta^*(\psi)$. Scoring is the \emph{outer loop}: a meta-objective $\mathcal{L}_{\mathrm{meta}}$ (defined below) rates how well $\vtheta^*(\psi)$ forgets and retains. We improve $\Phi$ by adjusting $\psi$ to lower that score. The catch is that changing $\psi$ also changes the model the inner loop returns, so $\psi$ and $\vtheta$ cannot be optimized separately; it is a \emph{bilevel} problem:
\begin{equation}
\vtheta^*(\psi) = \operatorname*{arg\,min}_{\vtheta} \mathcal{L}_{\mathrm{inner}}(\vtheta;\, \psi),
\quad
\min_{\psi}\ \mathcal{L}_{\mathrm{meta}}\big(\vtheta^*(\psi)\big).
\label{eq:bilevel}
\end{equation}
The score $\mathcal{L}_{\mathrm{meta}}$ depends on $\psi$ only through this model, never on $\Phi$ directly. In practice the inner loop is not run to convergence: $\vtheta^*(\psi)$ is the model after a fixed, short number of unlearning steps.

\mypar{The meta-objective}
\looseness=-1 $\mathcal{L}_{\mathrm{meta}}$ scores the two sides of unlearning separately, by what the unlearned model generates. For every forget target $c$ in the set $\mathcal{C}$, the model generates images for the target's prompt, and we take three measurements. The first is their SSCD similarity $M_c$ to the memorized image, our forgetting metric (Section~\ref{sec:setup}). The second is their CLIP-IQA quality $Q_c$. The third is their fidelity $F_c$, i.e., the CLIP similarity to the retain images of the prompt. Together,
\begin{equation}
\begin{split}
\mathcal{L}_{\mathrm{meta}}(\vtheta) = \frac{1}{|\mathcal{C}|} \sum_{c \in \mathcal{C}} \Big[\, & \mu\, \max(M_c - \tau_c,\, 0) + (1 - Q_c) \\
& + \nu\, \max(F_c^{\mathrm{r}} - F_c,\, 0) + \rho\, R_c \Big].
\end{split}
\label{eq:meta}
\end{equation}
The first term is the forget side. Stronger forgetting is not uniformly beneficial: once a sample has been erased, further minimizing the forget term degrades generation quality by corrupting the retain distribution. So we forget only down to a floor and then stop: the term is zero once $M_c$ is below a floor $\tau_c$ and grows above it. The floor is $0.1$, the similarity of unrelated images on this benchmark, or the closest level that the retain images of the target themselves reach. The second and third terms are the retain side: the images should look good, and they should look like what the prompt gives when nothing is memorized, where $F_c^{\mathrm{r}}$ is the fidelity that the retain images have among themselves. The last term, $R_c \in \{0, 1\}$, flags a target for which any generated image is still a replication (SSCD above $0.5$). We use $\mu = 4$, $\nu = 2$, and $\rho = 1$.

\mypar{Searching without gradients}
\looseness=-1 The outer loop needs to know how the score changes when $\psi$ changes. The obvious route, differentiating through the inner loop, is closed here: the score passes through image generation and through metrics that have no gradient. We therefore measure the change directly. One evaluation of $\mathcal{L}_{\mathrm{meta}}$ is one unlearning run followed by scoring its generations. We start from the squared-$L^2$ loss, $\psi = 0$. A step of the search moves each parameter up and down in turn to obtain a finite-difference estimate of the gradient, and then moves $\psi$ along the descent direction to the best of a few step sizes. Two runs with the same $\psi$ can differ in their score by as much as a single parameter changes it, so we accept no candidate on one run: we run the best candidates of a step again with further seeds and keep the one with the lowest mean score. We took three such steps. The first two used the meta-objective without its fidelity term, and showed us that quality and forgetting, as measured, can be improved at the expense of fidelity to the prompt (Section~\ref{sec:results}). The third used the full objective~\eqref{eq:meta}.

\mypar{Cross-fitting}
Choosing $\Phi$ on the very images on which BEND is then evaluated would be tuning on the test set. We therefore split the forget targets at random into two halves, find one potential on each half, and unlearn every target under the potential found on the \emph{other} half. No image contributes to the loss it is unlearned with.

\mypar{Joint multi-target unlearning}
\looseness=-1 When the forget set has more than one element, we still train a \emph{single} model, cycling the inner loop through them in round-robin, each with its own prompt and retain images. Targets differ in how stubbornly they are memorized, so under one shared pressure the easy targets over-forget, destroying quality, before the hard ones are erased; prior work therefore focused on training one model per image. BEND instead checks every target at regular intervals during unlearning, with the same forgetting metric as above. A target that has reached a floor and shows no replication is retired from the forget term. It keeps a share of retain steps so that it is not disturbed by the others, and returns if it drifts back above the floor. Nothing in this procedure depends on the loss, so it can also be run with ReTrack's loss and target; this gives the baseline ReTrack-joint of Section~\ref{sec:experiments}. It differs from BEND only in the two components this section introduced.

\mypar{Computational cost}
\looseness=-1 BEND adds nothing at inference: its loss is used only during unlearning, so the unlearned model is an ordinary diffusion U-Net with unchanged sampling cost. During unlearning the inner loop \emph{is} ReTrack with a different target and a weighted loss; unlearning all $48$ benchmark images in one model takes $39$ to $49$ minutes on one RTX~4090, including the checks. Finding $\Phi$ is the expensive part, about $35$ GPU-hours per potential, but it is paid once: the potential is applied to images it was not searched on. The unlearning runs within a step of the search are independent of each other, so the search is embarrassingly parallel: on eight GPUs, one potential takes about five hours.

\section{Experiments}
\label{sec:experiments}

We compare BEND against prior work, assessing whether searching for the loss can deliver better unlearning in practice: more thorough forgetting at higher retained generation quality. Stable Diffusion, the one production-scale real-world setting, carries the most weight, and three further benchmarks complete the ReTrack protocol.

\subsection{Setup}
\label{sec:setup}

We evaluate on the four benchmarks of the ReTrack protocol, against ReTrack. The models are DDPMs throughout, the Gaussian-affine special case of flow matching \citep[\S10]{lipman2024fmguide}, matching ReTrack's setting. The code for all experiments will be released upon publication under the GPL-2.0 license.

\mypar{Benchmarks}
\looseness=-1 {\em Stable Diffusion}: Stable Diffusion v1.4 \citep{rombach2022ldm} on the standard SISS benchmark because data unlearning requires \emph{verifiably} memorized training pairs. Its public LAION training set underlies the only extensively documented memorization corpus \citep{carlini2023extracting, somepalli2023diffusion}, and it is the testbed on which all published baselines report. ReTrack's Stable Diffusion evaluation, its only one, unlearns each of the corpus's $48$ images independently and reports corpus-averaged quality and forgetting curves without per-image numbers. For each of these images the benchmark provides generations of the pretrained model for the image's prompt, separated by clustering into those that replicate the memorized image and those that do not; the latter, $66$ per target on average, are the retain images of the target. We evaluate on the full corpus of all $N = 48$ verifiably memorized images and compare BEND, a \emph{single} model that forgets all targets simultaneously, with three ReTrack systems: a one-model-per-target \emph{fleet} (ReTrack-fleet), sequential unlearning (ReTrack-cascade: each request applied to the model the previous request produced), and ReTrack-joint, a single model trained with ReTrack's loss and target under exactly the joint procedure of BEND.

\emph{CelebA-HQ} ($256 \times 256$): ten forget identities, one fine-tuned model per identity for every method. \emph{MNIST T-Shirt}: forget an injected T-shirt class; ten random seeds. \emph{CIFAR-10}: ten individual forget images. These three models have no prompts and therefore no retain images per target, so BEND redirects to the mean of the $k = 10$ nearest retain images, and the potential is found for each dataset with the search of Section~\ref{sec:method} (supplementary material).

\mypar{Metrics}
\looseness=-1 Forgetting is measured by SSCD \citep{pizzi2022sscd}, the self-supervised copy-detection similarity to the forgotten image (lower is better): we score prompt-conditioned generations against the memorized training image, for the memorized prompt and for a paraphrase of it (the modified prompt). Generation quality is measured by CLIP-IQA \citep{wang2023clipiqa} (higher is better), on both prompts and on twelve unrelated prompts. CLIP-IQA judges an image without reference to the prompt. We therefore add two measures of prompt fidelity that the benchmark does not have: CLIP-T, the CLIP similarity between a generation and its prompt, and CLIP-I, the CLIP similarity between a generation and the retain images of the prompt. Jointly trained systems are run with three seeds; we report means and standard deviations over seeds. To compare two systems, we report the mean over the $48$ targets of the per-target difference of their seed-averaged values, and call the difference significant if its bootstrap $95\%$ confidence interval over the targets \citep{efron1993bootstrap} excludes zero, i.e., if it is not an artifact of the particular choice of targets. The intervals are listed in the supplementary material. The three unconditional benchmarks use the metrics of ReTrack's protocol (NLL of the forgotten image, SSCD of its reconstruction, FID \citep{heusel2017fid}, and IS \citep{salimans2016improved}), described in the supplementary material.

\mypar{Training}
\looseness=-1 BEND and ReTrack-joint share the procedure of Section~\ref{sec:method}: $24$ optimizer steps of $64$ samples each, $\lambda = 0.5$, and the same checks every eight steps with the same rule for retiring targets. The potential of BEND is found on two halves of $24$ targets and cross-fitted. ReTrack-fleet and ReTrack-cascade use the released settings of ReTrack; details are in the supplementary material.

\begin{figure}[t]
	\centering
	\includegraphics[width=\columnwidth]{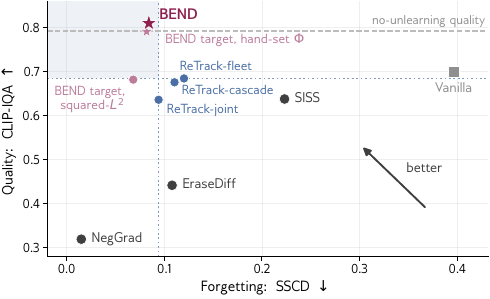}
	\caption{Quality versus forgetting on Stable Diffusion (system means over the $48$ targets). Vanilla, NegGrad, EraseDiff, and SISS train one model per target. BEND, one model for all targets, lies inside the shaded region of outcomes that dominate every ReTrack system, at the quality of the model before unlearning (dashed line). The two lighter marks are BEND with its searched potential replaced by a hand-set one (star) and by the squared-$L^2$ loss (circle).}
	\label{fig:sd-scatter}
\end{figure}

\begin{table*}[t]
	\centering
	\renewcommand{\arraystretch}{0.97}
	\small
	\setlength{\tabcolsep}{3.2pt}
	\begin{tabular}{llrrrrrrr}
		\toprule
		& & \multicolumn{2}{c}{Memorized prompt} & \multicolumn{2}{c}{Modified prompt} & \multicolumn{1}{c}{Unrelated} & \multicolumn{2}{c}{Prompt fidelity} \\
		\cmidrule(lr){3-4} \cmidrule(lr){5-6} \cmidrule(lr){7-7} \cmidrule(lr){8-9}
		& Method & \multicolumn{1}{c}{CLIP-IQA $\uparrow$} & \multicolumn{1}{c}{SSCD $\downarrow$} & \multicolumn{1}{c}{CLIP-IQA $\uparrow$} & \multicolumn{1}{c}{SSCD $\downarrow$} & \multicolumn{1}{c}{CLIP-IQA $\uparrow$} & \multicolumn{1}{c}{CLIP-T $\uparrow$} & \multicolumn{1}{c}{CLIP-I $\uparrow$} \\
		\midrule
		& Pretrained$^{\dagger}$ & $0.792$ & $0.770$ & $0.788$ & $0.548$ & $0.904$ & $0.287$ & $0.723$ \\
		\midrule
		$48$ models & ReTrack-fleet & $0.685 \pm 0.006$ & $0.120 \pm 0.003$ & $0.743$ & $0.088$ & $0.897$ & $0.259$ & $0.689$ \\
		\midrule
		One model & ReTrack-cascade & $0.676 \pm 0.019$ & $0.110 \pm 0.016$ & $0.716$ & $0.120$ & $0.872$ & $0.241$ & $0.651$ \\
		& ReTrack-joint & $0.636 \pm 0.044$ & $0.094 \pm 0.003$ & $0.746$ & $0.081$ & $0.896$ & $0.248$ & $0.658$ \\
		& BEND (ours) & $0.811 \pm 0.002$ & $0.084 \pm 0.004$ & $0.844$ & $0.098$ & $0.922$ & $0.246$ & $0.651$ \\
		\midrule
		Ablations & ReTrack target, hand-set $\Phi$ & $0.771 \pm 0.032$ & $0.172 \pm 0.041$ & $0.843$ & $0.123$ & $0.921$ & $0.251$ & $0.661$ \\
		& BEND target, squared-$L^2$ & $0.682 \pm 0.026$ & $0.068 \pm 0.003$ & $0.746$ & $0.072$ & $0.901$ & $0.247$ & $0.653$ \\
		& BEND target, hand-set $\Phi$ & $0.791 \pm 0.013$ & $0.082 \pm 0.004$ & $0.827$ & $0.089$ & $0.918$ & $0.233$ & $0.631$ \\
		& BEND, no fidelity term & $0.803 \pm 0.015$ & $0.063 \pm 0.005$ & $0.835$ & $0.082$ & $0.922$ & $0.231$ & $0.618$ \\
		\bottomrule
	\end{tabular}
	\caption{Stable Diffusion, all $N = 48$ memorized images (\emph{-fleet}: one model per target; \emph{-cascade}: one model, sequential; \emph{-joint}: one model for all targets, trained with the procedure of BEND). $\pm$: standard deviation over three seeds (the fidelity of the fleet is from one of them). \emph{Ablations}: the joint procedure of BEND with the target or the loss exchanged; \emph{no fidelity term}: the potential found with the meta-objective without its fidelity term. \emph{Unrelated}: CLIP-IQA on twelve unrelated prompts. $^{\dagger}$no unlearning. Per-target distributions are in the supplementary material.}
	\label{tab:sd}
\end{table*}

\subsection{Results}
\label{sec:results}

\mypar{Stable Diffusion, \boldmath$N = 48$}
\looseness=-1 Table~\ref{tab:sd}, depicted in Fig.~\ref{fig:sd-scatter}, reports the comparison. The three ReTrack systems end close to each other, although they differ greatly in cost. Forty-eight models, a sequence of forty-eight updates, or one joint run all leave SSCD between $0.094$ and $0.120$ and CLIP-IQA between $0.64$ and $0.69$, well below the $0.79$ of the model before unlearning. For $8$ to $18$ of the $48$ prompts, depending on the seed, the generations of ReTrack-joint are visibly degraded (CLIP-IQA below $0.5$). BEND, a \emph{single} model trained to forget all $48$ images at once, holds CLIP-IQA at $0.811$, the level of the pretrained model, and at most one prompt per seed is degraded. Against ReTrack-joint, which shares its entire training procedure, the paired difference per target is a significant $+0.174$ in CLIP-IQA, with higher quality on $45$ of the $48$ targets. BEND also forgets more than ReTrack-fleet and ReTrack-cascade, by $0.036$ and $0.026$ in SSCD; both differences are significant. Against ReTrack-joint the difference is $0.010$, which is not significant. Under the modified prompt ReTrack-joint and ReTrack-fleet end lower than BEND (SSCD $0.081$ and $0.088$ against $0.098$); BEND, which stops forgetting at the floor of $0.1$, stays below it. Prompt fidelity, which CLIP-IQA cannot see (last two columns), suffers under \emph{every} method: the similarity to the prompt falls from $0.287$ for the pretrained model to $0.23$--$0.26$. BEND matches ReTrack-joint in both measures, and the fleet, which spends one model on every target, keeps the most; details are in the supplementary material. One of the three seeds leaves a single replicated generation for one target.

\mypar{Target and loss}
\looseness=-1 BEND differs from ReTrack-joint in two components, and the ablations in Table~\ref{tab:sd} exchange them one at a time. The target decides how much is forgotten. With the squared-$L^2$ loss kept, moving from ReTrack's redirect image to the mean of the retain images removes more than a quarter of the remaining similarity (SSCD $0.094 \to 0.068$). But quality stays where it was ($0.636 \to 0.682$), i.e., the model forgets and blurs. The loss decides what the result looks like: a potential that ignores the finest scales raises CLIP-IQA by $0.11$ to $0.14$ under either target. Applied to ReTrack's own target, however, it gives the weakest forgetting among the joint systems (SSCD $0.172$, with replicated generations for one to six targets, depending on the seed), since the model is still redirected to the near-copy of Fig.~\ref{fig:redirect}. Only the two changes together forget more than ReTrack \emph{and} keep the quality. Two further ablations, on the number of retain images behind the target and on the retain term, are in the supplementary material.

\begin{figure}[t]
	\centering
	\includegraphics[width=\columnwidth]{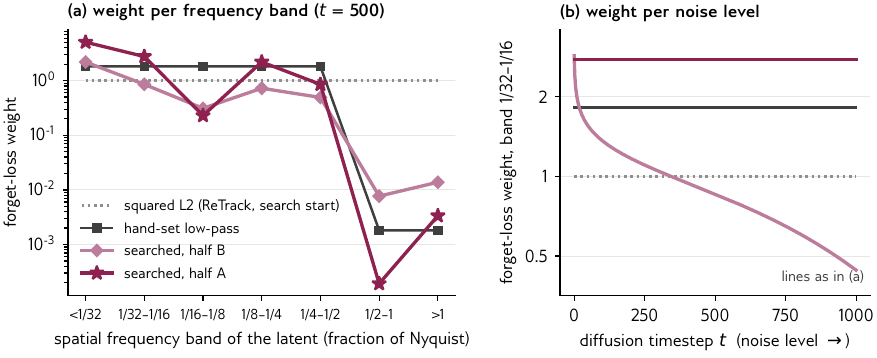}
	\caption{The loss found by the search. Weight of the forget term (a) per frequency band and (b) per noise level, for the squared-$L^2$ loss where the search starts, a hand-set low-pass, and the potentials found on the two halves of the targets.}
	\label{fig:gates}
\end{figure}

\mypar{What the search finds}
\looseness=-1 Fig.~\ref{fig:gates} shows the potentials it finds. Starting from the squared-$L^2$ loss, the first step closes the two finest frequency bands on both halves of the targets. It lowers their weight by one to two orders of magnitude against the coarsest band, and the second step sharpens this shape. A loss that acts on coarse scales only is what \citet{park2025frequency} propose by design; here it emerges from the search, without any prior on the shape, and on both halves independently. These two steps used the meta-objective without its fidelity term. The resulting potential forgets the most of all systems (SSCD $0.063$, last row of Table~\ref{tab:sd}), but at the lowest fidelity. The third step, with the fidelity term, changes a single parameter, the same on both halves, and it is the one change that stands clear of the variation between runs: the exponent $p$, which the first steps had used to move weight toward high noise levels, is lowered by the same amount on both halves, to $0$ on one and $-0.15$ on the other. This restores fidelity (CLIP-I $0.618 \to 0.651$) and gives up forgetting only down to the floor that the meta-objective asks for (SSCD $0.063 \to 0.084$). A hand-set potential with the two finest bands closed (Table~\ref{tab:sd}) shares most of the gain, so the shape is the right one. At equal forgetting (SSCD $0.084$ against $0.082$), the searched potential is better in quality, by $0.019$ in CLIP-IQA, and in fidelity, by $0.013$ in CLIP-T and $0.020$ in CLIP-I; all three differences are significant, the first one narrowly.

\mypar{Unconditional benchmarks}
\looseness=-1 On the three benchmarks of ReTrack, where one image is forgotten at a time, BEND matches or exceeds the state of the art. Both methods remove the MNIST T-shirt from all generations, with FID within the variation between seeds; on CIFAR-10 BEND forgets more by both measures at the same quality, and on CelebA-HQ it is better in all three measures of the benchmark, against ReTrack run under the same conditions (NLL $2.09$ vs $1.52$, SSCD $0.296$ vs $0.393$, FID $23.8$ vs $25.4$; tables in the supplementary material).

\section{Conclusion}
\label{sec:conclusion}
\looseness=-1 Data unlearning has treated the training loss as fixed and searched only over how the model is fine-tuned. We showed that the choice of loss is a fundamental lever: valid diffusion losses are exactly the Bregman family, and prior work has used the squared-$L^2$, at most masked by hand. BEND searches for the loss instead, fitting a Bregman potential to the unlearning task through a bilevel loop that judges it by what the unlearned model generates. It also redirects a memorized image to the mean of its retain images instead of ReTrack's single near-copy. This gives a single model that erases all $48$ memorized images of the Stable Diffusion benchmark at least as thoroughly as per-target, sequential, and jointly trained ReTrack, at the quality of the model before unlearning, and it outperforms ReTrack on CelebA-HQ and matches or exceeds it on MNIST and CIFAR-10. Like every method we measured, BEND loses some fidelity to the prompt, about as much as jointly trained ReTrack. Two limitations remain: the search scores a loss with the metrics we also evaluate and runs per benchmark, so transfer to held-out measures and across datasets is untested; and redirecting to the mean of the model's own generations suppresses the memorized content, which a model retrained with its near-duplicates might still produce.

{
    \small
    \bibliographystyle{ieeenat_fullname}
    \bibliography{main}
}

\clearpage
\makeatletter\newwrite\arxiv@suppaux\immediate\write\@mainaux{\string\arxiv@suppaux@input}
\immediate\openout\arxiv@suppaux=supp.aux\immediate\write\arxiv@suppaux{\relax}
\immediate\write\arxiv@suppaux{\string\bibstyle{ieeenat_fullname}}
\immediate\write\arxiv@suppaux{\string\bibdata{supp}}
\let\@auxout\arxiv@suppaux\makeatother
\renewcommand{\topfraction}{0.95}
\renewcommand{\bottomfraction}{0.6}
\renewcommand{\textfraction}{0.05}
\renewcommand{\floatpagefraction}{0.9}
\setcounter{topnumber}{3}
\setcounter{totalnumber}{5}
\setcounter{dbltopnumber}{3}
\renewcommand{\dbltopfraction}{0.95}
\renewcommand{\dblfloatpagefraction}{0.9}

\appendix
\addtocontents{toc}{\protect\setcounter{tocdepth}{2}}
\twocolumn[{\centering\Large\textbf{Supplementary Material}\par\vspace{1.5em}}]

\tableofcontents

\section{Why Bregman, and only Bregman: details}
\label{app:bregman}

This appendix expands the claim the main paper rests on: replacing the squared-$L^2$ per-sample loss is \emph{safe} exactly when the replacement is a Bregman divergence, and \emph{unsafe} for every loss outside that family. We spell out both directions and then locate what does change inside the family, which is the design space BEND searches.

\mypar{Two objectives}
Let $\ell^\Phi_t(\vtheta;\, \va) = D_\Phi(\vepsilon_t(\va), \vepsilon_{\vtheta}(\vx_t, t))$ be the Bregman generalization of the per-sample loss $\ell_t$, and $\bar{\vepsilon}_t(\vx_t) = \E[\vepsilon_t(\va) \mid \vx_t]$ the noise field marginalized over the dataset. The objective one would like to optimize is the \emph{marginal} loss
\begin{equation}
\ell_{\mathrm{marg}}(\vtheta) = \E_{t, \vx_t}\, D_\Phi(\bar{\vepsilon}_t(\vx_t), \vepsilon_{\vtheta}(\vx_t, t)),
\label{eq:marg-loss}
\end{equation}
which is intractable: computing $\bar{\vepsilon}_t(\vx_t)$ requires integrating against every training sample. The tractable surrogate is the \emph{conditional} loss
\begin{equation}
\ell_{\mathrm{cond}}(\vtheta) = \E_{t, \va, \vx_t \sim q_t(\cdot \mid \va)}\, \ell^\Phi_t(\vtheta;\, \va),
\label{eq:cond-loss}
\end{equation}
which only ever needs the noise target of a single sample. Validity means the swap is free: descending \eqref{eq:cond-loss} descends \eqref{eq:marg-loss}.

\addcontentsline{toc}{subsection}{Sufficiency}
\mypar{Sufficiency: the two gradients coincide for every Bregman divergence}
The computation is short enough to give in full. Fix $t$ and $\vx_t$, write $\vv = \vepsilon_{\vtheta}(\vx_t, t)$ for the model output, and take the conditional expectation of the Bregman loss over $\va \mid \vx_t$:
\begin{multline}
\E_{\va \mid \vx_t} D_\Phi(\vepsilon_t(\va), \vv) \\
= \E[\Phi(\vepsilon_t(\va))] - \Phi(\vv) - \langle \nabla \Phi(\vv),\, \bar{\vepsilon}_t - \vv \rangle,
\label{eq:cond-exp}
\end{multline}
using only linearity of expectation in the first argument. Comparing with
\begin{equation}
D_\Phi(\bar{\vepsilon}_t, \vv)
= \Phi(\bar{\vepsilon}_t) - \Phi(\vv) - \langle \nabla \Phi(\vv),\, \bar{\vepsilon}_t - \vv \rangle,
\end{equation}
the two differ by $\E[\Phi(\vepsilon_t(\va))] - \Phi(\bar{\vepsilon}_t)$, which does not depend on $\vv$ and hence not on $\vtheta$. The pointwise gap between \eqref{eq:cond-loss} and \eqref{eq:marg-loss} is therefore a constant in the parameters, so
\begin{equation}
\nabla_{\vtheta} \ell_{\mathrm{cond}}(\vtheta) = \nabla_{\vtheta} \ell_{\mathrm{marg}}(\vtheta).
\label{eq:grad-identity}
\end{equation}
The mechanism is that $\nabla_{\vv} D_\Phi(\vu, \vv)$ is affine in its first argument, so expectations can be pulled through the gradient; \citet[Eqs.~(4.20)--(4.21)]{S-lipman2024fmguide} state exactly this identity for flow matching, and their Theorem~4 and Proposition~1 build on it, of which diffusion training is the Gaussian-affine special case.

\addcontentsline{toc}{subsection}{Necessity}
\mypar{Necessity: outside the family the swap breaks}
Necessity is due to \citet[Theorems~3 and~4]{S-banerjee2005bregman} (the scalar and vector cases, the latter under mild smoothness): if a non-negative loss $F$ with $F(\vx, \vx) = 0$ has the conditional mean as the unique minimizer of $\vv \mapsto \E\,F(\vy, \vv)$ for \emph{every} distribution of $\vy$, then $F$ is a Bregman divergence. The proof idea is instructive: optimality on all two-point distributions already forces a functional equation on $F$'s derivatives whose smooth solutions are exactly the Bregman forms. Contrapositively, for any loss outside the family there exists a data distribution on which the minimizer of the conditional surrogate is \emph{not} $\bar{\vepsilon}_t$: descending the tractable loss then actively pulls the model away from the score, and no amount of training fixes it. This is why BEND constrains its search to the Bregman family by construction rather than penalizing violations.

\mypar{The mean of the targets}
The computation in~\eqref{eq:cond-exp} uses nothing about where the random target comes from, and the main paper relies on it twice more. First, ReTrack's forget term is an expected squared-$L^2$ loss over retain images drawn with the weights $w_t(\vx_t \mid \va_r)$; by the same argument it equals, up to a constant, the loss against the single target $\sum_r w_t(\vx_t \mid \va_r)\,\vepsilon_t(\va_r)$, which is the noise target of the weighted average image. This is the redirect image of Section~4. Second, regressing onto the mean target of the retain images, as BEND does, has the same gradient as drawing one retain image at random for every noisy sample, for every Bregman loss; the mean only removes the variance of the draw.

\mypar{What does change inside the family}
Validity fixes the optimum, not the trajectory. Differentiating the Bregman divergence in its second argument gives
\begin{equation}
\nabla_{\vv} D_\Phi(\vu, \vv) = \nabla^2 \Phi(\vv)\, (\vv - \vu),
\end{equation}
so the Hessian of the potential acts as a metric on the error: it reweights error directions, rescales per-sample gradient magnitudes, and thereby changes gradient noise, conditioning, and every finite-step property of training. Unlearning lives entirely in this finite-step regime: budgets are short, the sampling is deliberately off-distribution (ReTrack's near-forget proposals), and runs stop after a few dozen steps rather than at convergence. In that regime the geometry $\nabla^2 \Phi$, not the shared population optimum, determines which directions of change reach the forget target first and what collateral movement the retain distribution suffers. That geometry is exactly the quantity BEND's search shapes. For the potential of the main paper the Hessian is constant and diagonal in the spatial Fourier basis, with the weights $w_\psi(t, b)$ on its diagonal. The squared-$L^2$ loss is the special case $\Phi(\vu) = \tfrac{1}{2}\|\vu\|^2$ with $\nabla^2 \Phi = \mI$, the one point of the family whose geometry is isotropic; in this notation $\ell_{\mathrm{cond}}$ reduces to the standard DDPM objective $\E\,\ell_t(\vtheta;\,\va)$.

\section{The redirection target: details}
\label{app:redirect}

\addcontentsline{toc}{subsection}{Why the weights collapse}
\mypar{Why the weights collapse}
For two retain images $\va_r, \va_{r'}$ and a noisy forget sample $\vx_t = \gamma_t \va_u + \sigma_t\vepsilon$, the ratio of ReTrack's weights is
\begin{multline}
\log\frac{w_t(\vx_t \mid \va_r)}{w_t(\vx_t \mid \va_{r'})} = \frac{\gamma_t^2}{2\sigma_t^2}\Big(\|\va_u - \va_{r'}\|^2 - \|\va_u - \va_r\|^2\Big) \\
+ \frac{\gamma_t}{\sigma_t}\,\langle \vepsilon,\, \va_r - \va_{r'} \rangle.
\end{multline}
The first term grows with the dimension $d$ of the latent and the second only with $\sqrt{d}$. This decides the outcome. For Stable Diffusion $d = 16{,}384$, and two retain images are never at the same distance from $\va_u$ to within the precision this would require. The first term is therefore far from zero unless the signal-to-noise ratio $\gamma_t^2/\sigma_t^2$ is of the order of $1/d$, which the schedule reaches only at its last steps. The weights over any finite set of images therefore concentrate on the image nearest to $\va_u$.

\mypar{Measurement}
For each of the $48$ targets we noise up to $16$ memorized latents at seven timesteps and compute ReTrack's weights over its $k = 10$ neighbors, and for comparison the same weights over all retain images of the target. Table~\ref{tab:redirect} reports the effective number of images $1/\sum_r w_r^2$ and the SSCD similarity of the redirect image to the memorized image, $\sum_r w_r\, s_r$ with $s_r$ the similarity of retain image $r$. The retain images as a whole have a mean similarity of $0.165$ to the memorized image, and the ten neighbors of $0.266$.

\begin{figure*}[t]
\centering
\includegraphics[width=0.95\textwidth]{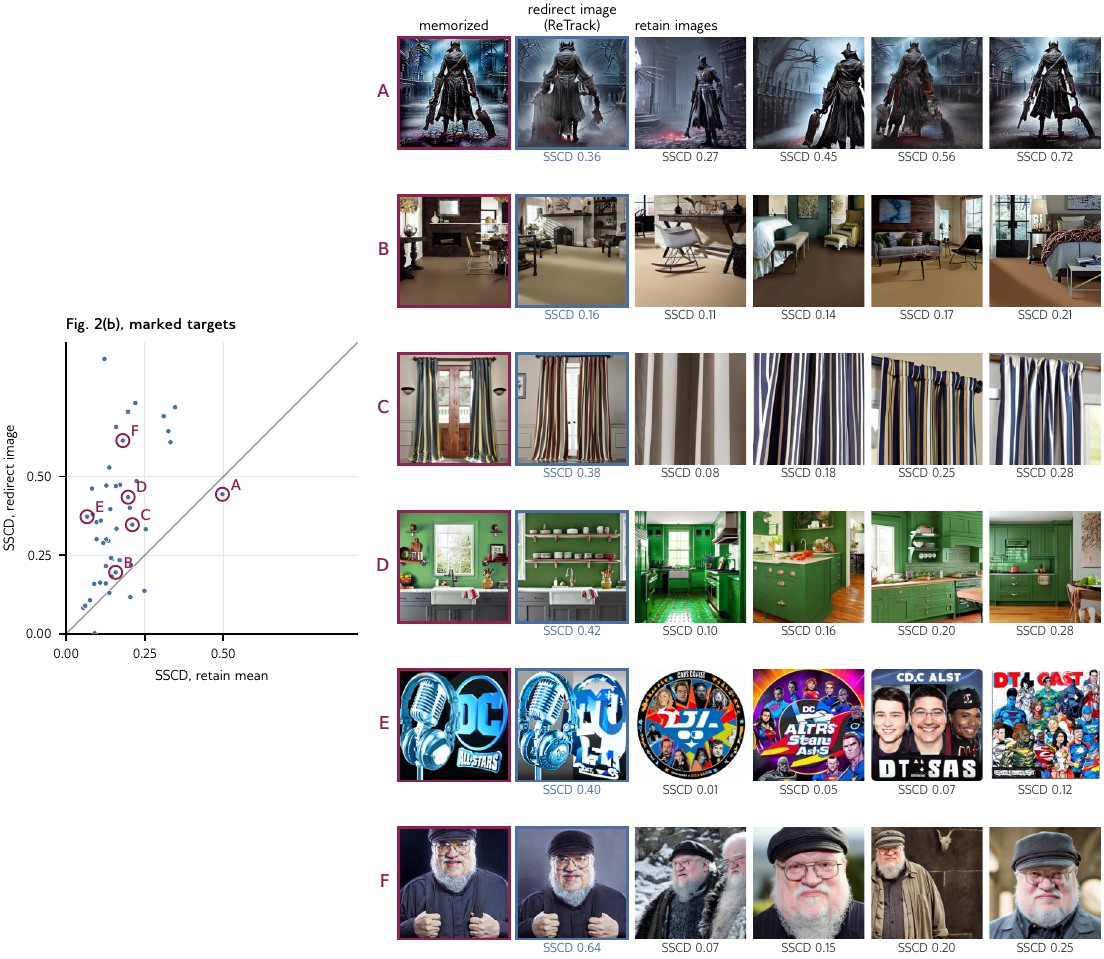}
\caption{Examples for Fig.~2(b) of the main paper. Left: the plot of Fig.~2(b), with six targets marked. Right: one row per marked target, with the memorized image, ReTrack's redirect image (the retain image onto which its weights collapse, here the one selected most often over the memorized generations of the target), and four retain images at evenly spaced quantiles of their similarity to the memorized image. Numbers: SSCD similarity to the memorized image. The point in Fig.~2(b) averages over the memorized generations and noise levels, so its height can differ slightly from the similarity of the redirect image shown.}
\label{fig:redirect-examples}
\end{figure*}

\addcontentsline{toc}{subsection}{Examples}
\mypar{Examples}
Fig.~\ref{fig:redirect-examples} shows six of the targets of Fig.~2(b) of the main paper, chosen at evenly spaced positions in the ranking of the targets by the height of their point above the diagonal. For five of them the redirect image is more similar to the memorized image than the retain images on average, and for three (D, E, and F) it is a near-copy that shows the same scene, object, or person in almost the same composition. For target~A the retain images themselves are close to the memorized image, and the redirect image lies below their average. In most cases the weights select the same retain image for (nearly) all memorized generations of a target: for $14$ to $16$ of $16$ in C to F, for $9$ of $16$ in B, and for $6$ of $12$ in A, where two images share the weight.

\begin{table}[t]
\centering
\small
\setlength{\tabcolsep}{4pt}
\begin{tabular}{rccc}
\toprule
 & \multicolumn{2}{c}{Effective number} & SSCD, \\
\cmidrule(lr){2-3}
$t$ & $k = 10$ & all retain & redirect \\
\midrule
$20$ & $1.00$ & $1.00$ & $0.359$ \\
$100$ & $1.00$ & $1.00$ & $0.361$ \\
$200$ & $1.00$ & $1.00$ & $0.362$ \\
$400$ & $1.00$ & $1.00$ & $0.363$ \\
$600$ & $1.01$ & $1.01$ & $0.366$ \\
$800$ & $1.05$ & $1.05$ & $0.366$ \\
$999$ & $1.39$ & $1.52$ & $0.324$ \\
\bottomrule
\end{tabular}
\caption{ReTrack's redirection on Stable Diffusion (means over the $48$ targets). The weights are a point mass at all but the last timesteps, with or without the truncation to $k$ neighbors.}
\label{tab:redirect}
\end{table}

\addcontentsline{toc}{subsection}{The tilted target}
\mypar{The tilted target}
The variant mentioned in Section~4 of the main paper reweights the retain images $\sS$ of a target away from the memorized image. We denote with $s(\va_r)$ the SSCD similarity of the retain image $\va_r$ to the memorized image and with $u$ the uniform distribution on $\sS$, which BEND uses as weighting. Among all distributions $q$ on $\sS$ whose mean similarity $\sum_r q(\va_r)\, s(\va_r)$ is at most a level $\kappa$, the one that minimizes the Kullback--Leibler divergence $\mathrm{KL}(q \,\|\, u)$ is
\begin{equation}
q_\beta(\va_r) \propto e^{-\beta\, s(\va_r)}, \qquad \beta \ge 0,
\end{equation}
where $\beta$ is the multiplier of the constraint, which is of course $0$ if the retain images already satisfy it. In words, a retain image loses weight exponentially with its similarity to the memorized image, and $\beta$ sets how fast. Since the Kullback--Leibler divergence is itself a Bregman divergence, generated by the negative entropy, $q_\beta$ is the Bregman projection of $u$ onto the constraint. Note that both the mean similarity and the effective number of images $1/\sum_r q_\beta(\va_r)^2$ decrease with $\beta$. We thus take the smallest $\beta$ that reaches the level, but stop increasing it where fewer than eight images would carry effective weight. Otherwise the target could collapse onto only few images, like the weights of ReTrack discussed above. We then redirect the forget term to the mean of the retain images under $q_\beta$ and draw the images of the retain term from $q_\beta$. For $\kappa = 0.10$, $11$ of the $48$ targets need no tilt and $5$ reach the limit of eight images before the level. On average the mean similarity of the tilted retain images is $0.098$ (against $0.165$ without tilt), and $45$ of $66$ images carry effective weight.

\section{The potential: details}
\label{app:potential}

\mypar{Bands and noise levels}
The prediction error lives on the $64 \times 64$ latent grid with four channels. The bands of~Eq.~(9) of the main paper are octaves of the radial spatial frequency with edges at $1/32$, $1/16$, $1/8$, $1/4$, $1/2$ and $1$ times the Nyquist frequency of an axis. The seventh band holds the corners of the frequency plane beyond it. The two finest bands hold $59\%$ and $22\%$ of all frequencies. The dependence on the noise level enters through the log signal-to-noise ratio only, Eq.~(10), so the potential has no free parameter per timestep.

\addcontentsline{toc}{subsection}{Normalization}
\mypar{Normalization}
The constant $c(\psi)$ fixes the total weight of the forget term: the weights average to one, their value for the squared-$L^2$ loss. In this average each timestep is weighted by its signal-to-noise ratio and each band by its share in the energy of the forget residual, a fixed constant that we measured once on the benchmark, before the search and independently of $\psi$. Without it, closing a band would also weaken the forget term as a whole, and the search could not tell the shape of the loss from its strength.

\addcontentsline{toc}{subsection}{The potentials}
\mypar{The potentials}
Table~\ref{tab:psi} lists the parameters after each step of the search. The hand-set low-pass used as an ablation has $g_b = 0$ for the five coarsest bands and closes the two finest (weight $10^{-3}$), with $p = 0$.

\begin{table}[t]
\centering
\small
\setlength{\tabcolsep}{1.9pt}
\begin{tabular}{llrrrrrrrr}
\toprule
 & Step & $g_0$ & $g_1$ & $g_2$ & $g_3$ & $g_4$ & $g_5$ & $g_6$ & $p$ \\
\midrule
Half A & 1 & $2.19$ & $1.78$ & $1.07$ & $1.63$ & $0.06$ & $-3.33$ & $-6.00$ & $0.30$ \\
 & 2 & $3.28$ & $2.67$ & $0.18$ & $2.44$ & $1.50$ & $-6.90$ & $-4.04$ & $0.30$ \\
 & 3 & $3.28$ & $2.67$ & $0.18$ & $2.44$ & $1.50$ & $-6.90$ & $-4.04$ & $0.00$ \\
\midrule
Half B & 1 & $0.89$ & $1.26$ & $-0.28$ & $0.86$ & $0.12$ & $-3.00$ & $-1.43$ & $0.00$ \\
 & 2 & $1.77$ & $0.83$ & $-0.18$ & $0.67$ & $0.28$ & $-3.88$ & $-3.29$ & $0.15$ \\
 & 3 & $1.77$ & $0.83$ & $-0.18$ & $0.67$ & $0.28$ & $-3.88$ & $-3.29$ & $-0.15$ \\
\bottomrule
\end{tabular}
\caption{Parameters $\psi$ found by the search (band gates from the coarsest band $g_0$ to the finest $g_6$, and the noise-level exponent $p$) on the two halves of the targets. The targets of one half are unlearned with the potential of the other. Gates and exponent act on the natural logarithm of the weights (Eq.~(10) of the main paper).}
\label{tab:psi}
\end{table}

\section{Training and evaluation protocol}
\label{app:protocol}

\mypar{Unlearning}
We fine-tune the U-Net of Stable Diffusion v1.4 with AdamW (learning rate $10^{-5}$, weight decay $10^{-2}$, constant schedule). One optimizer step accumulates $64$ samples, taken round-robin over the active targets; each sample contributes a forget term and a retain term with $\lambda = 0.5$. Runs have $32$ steps with the model stored after $16$, $24$, and $32$ steps; all results are reported after $24$ steps unless stated otherwise, a length we fixed from the in-run checks before any held-out evaluation.

\addcontentsline{toc}{subsection}{Checks during unlearning}
\mypar{Checks during unlearning}
Every eight steps the model generates eight images per target. A target is retired from the forget term when its mean SSCD is at most its retirement level, half the mean SSCD of its retain images to the memorized image but at least $0.03$ (this level is not the floor $\tau_c$ of the meta-objective), and none of its generations exceeds $0.5$. A retired target keeps a quarter of its samples on the retain term and returns to the forget term if its SSCD rises more than $0.03$ above this level or one of its generations exceeds $0.5$. BEND, ReTrack-joint, and all ablations use these checks unchanged.

\addcontentsline{toc}{subsection}{Baselines}
\mypar{Baselines}
ReTrack-fleet trains one model per target with the released implementation and settings of ReTrack ($30$ steps, $k = 10$, $\lambda = 0.5$); ReTrack-cascade applies the same per-target update to one model, target after target. ReTrack-joint uses ReTrack's forget term, with its $k = 10$ neighbors, importance weights, and squared-$L^2$ loss, inside the joint procedure above. We ran it in two versions. The first has the fixed length of $24$ steps that BEND uses and is reported in the main paper. The second stops as soon as the mean in-run SSCD reaches the level of unrelated images (Table~\ref{tab:full}). The two agree. Vanilla, NegGrad, EraseDiff, and SISS in Fig.~3 of the main paper train one model per target like ReTrack-fleet, with $30$ steps of $64$ samples and the optimizer above; SISS uses $\lambda = 0.5$ and a rescaling factor of $750$, and EraseDiff $\eta = 10^{-2}$. Table~\ref{tab:sd-all} adds them to Table~1 of the main paper.

\begin{table*}[t]
\centering
\renewcommand{\arraystretch}{0.97}
\small
\setlength{\tabcolsep}{3.2pt}
\begin{tabular}{llrrrrrrr}
\toprule
& & \multicolumn{2}{c}{Memorized prompt} & \multicolumn{2}{c}{Modified prompt} & \multicolumn{1}{c}{Unrelated} & \multicolumn{2}{c}{Prompt fidelity} \\
\cmidrule(lr){3-4} \cmidrule(lr){5-6} \cmidrule(lr){7-7} \cmidrule(lr){8-9}
& Method & \multicolumn{1}{c}{CLIP-IQA $\uparrow$} & \multicolumn{1}{c}{SSCD $\downarrow$} & \multicolumn{1}{c}{CLIP-IQA $\uparrow$} & \multicolumn{1}{c}{SSCD $\downarrow$} & \multicolumn{1}{c}{CLIP-IQA $\uparrow$} & \multicolumn{1}{c}{CLIP-T $\uparrow$} & \multicolumn{1}{c}{CLIP-I $\uparrow$} \\
\midrule
& Pretrained$^{\dagger}$ & $0.792$ & $0.770$ & $0.788$ & $0.548$ & $0.904$ & $0.287$ & $0.723$ \\
\midrule
$48$ models & ReTrack-fleet & $0.685 \pm 0.006$ & $0.120 \pm 0.003$ & $0.743$ & $0.088$ & $0.897$ & $0.259$ & $0.689$ \\
& Vanilla & $0.700$ & $0.396$ & $0.706$ & $0.204$ & $0.903$ & $0.276$ & $0.718$ \\
& NegGrad & $0.319$ & $0.015$ & $0.335$ & $0.009$ & $0.837$ & $0.129$ & $0.476$ \\
& EraseDiff & $0.441$ & $0.108$ & $0.597$ & $0.133$ & $0.891$ & $0.216$ & $0.615$ \\
& SISS & $0.638$ & $0.223$ & $0.696$ & $0.160$ & $0.902$ & $0.263$ & $0.707$ \\
\midrule
One model & ReTrack-cascade & $0.676 \pm 0.019$ & $0.110 \pm 0.016$ & $0.716$ & $0.120$ & $0.872$ & $0.241$ & $0.651$ \\
& ReTrack-joint & $0.636 \pm 0.044$ & $0.094 \pm 0.003$ & $0.746$ & $0.081$ & $0.896$ & $0.248$ & $0.658$ \\
& BEND (ours) & $0.811 \pm 0.002$ & $0.084 \pm 0.004$ & $0.844$ & $0.098$ & $0.922$ & $0.246$ & $0.651$ \\
\midrule
Ablations & ReTrack target, hand-set $\Phi$ & $0.771 \pm 0.032$ & $0.172 \pm 0.041$ & $0.843$ & $0.123$ & $0.921$ & $0.251$ & $0.661$ \\
& BEND target, squared-$L^2$ & $0.682 \pm 0.026$ & $0.068 \pm 0.003$ & $0.746$ & $0.072$ & $0.901$ & $0.247$ & $0.653$ \\
& BEND target, hand-set $\Phi$ & $0.791 \pm 0.013$ & $0.082 \pm 0.004$ & $0.827$ & $0.089$ & $0.918$ & $0.233$ & $0.631$ \\
& BEND, no fidelity term & $0.803 \pm 0.015$ & $0.063 \pm 0.005$ & $0.835$ & $0.082$ & $0.922$ & $0.231$ & $0.618$ \\
\bottomrule
\end{tabular}
\caption{Table~1 of the main paper with all baselines of the comparison of ReTrack. Vanilla, NegGrad, EraseDiff, and SISS train one model per target like ReTrack-fleet (one seed). For $26$ (Vanilla, NegGrad), $32$ (EraseDiff), and $33$ (SISS) of their $48$ models, the unrelated prompts and the prompt fidelity are scored on a retrained model of the same recipe, since the model behind the held-out numbers was not kept. All other entries and the notation are those of Table~1 of the main paper. The comparisons in the main paper (e.g., among all systems or the joint systems) refer to the systems of its Table~1.}
\label{tab:sd-all}
\end{table*}

\addcontentsline{toc}{subsection}{Evaluation}
\mypar{Evaluation}
Unless stated otherwise, all systems are evaluated with the sampler of the benchmark code, a DDPM scheduler with $50$ steps and the default guidance scale of the pipeline. SSCD uses the \texttt{sscd\_disc\_mixup} model at resolution $288$ and CLIP-IQA the \texttt{torchmetrics} implementation with its default prompts. Held-out generations use generator seeds that are disjoint from those of the in-run checks. We generate $64$ images per prompt for SSCD and CLIP-IQA and four images for each of twelve unrelated prompts. For CLIP-T and CLIP-I (CLIP ViT-L/14) we generate $16$ further images per prompt, with identical noise for every system. Intervals for the difference of two systems are bootstrap intervals over the $48$ targets of the per-target differences of seed-averaged values \citep{S-efron1993bootstrap}: the targets are resampled with replacement $4000$ times, and the interval spans the $2.5$th to the $97.5$th percentile of the resampled means. A difference is called significant if its interval excludes zero. Table~\ref{tab:paired} lists the intervals of the comparisons in the main paper.

\section{The search: details}
\label{app:learning}

Algorithm~\ref{alg:bend} lists the full procedure of BEND, and the paragraphs below give the details of its outer loop.

\begin{algorithm}[t]
\caption{BEND: bilevel unlearning}
\label{alg:bend}
\begin{algorithmic}[1]
\Require pretrained $\vtheta_0$; forget targets $\mathcal{C}$ with retain images; inner steps $S$; search steps $K$
\State $\psi \gets 0$ \Comment{squared-$L^2$ loss}
\For{$k = 1, \dots, K$} \Comment{outer loop: improve $\Phi$}
    \For{each candidate $\psi'$ around $\psi$}
        \State $\vtheta \gets \vtheta_0$
        \For{$s = 1, \dots, S$} \Comment{inner loop: unlearn}
            \State pick active target (round-robin); draw batches
            \State $\vtheta \gets \vtheta - \alpha\,\nabla_{\vtheta}\mathcal{L}_{\mathrm{inner}}(\vtheta;\, \psi')$ \Comment{Eqs.~(7),~(8) of the main paper}
            \State at intervals: retire or re-activate targets
        \EndFor
        \State score $\mathcal{L}_{\mathrm{meta}}(\vtheta)$ on generated images \Comment{Eq.~(12) of the main paper}
    \EndFor
    \State $\psi \gets$ best candidate, confirmed over seeds
\EndFor
\Ensure the model unlearned under the final $\psi$
\end{algorithmic}
\end{algorithm}

\mypar{Halves}
The $48$ targets are split once, at random, into two halves of $24$. A potential is found on each half by running the search with the joint procedure on its $24$ targets, and is applied in the $48$-target model to the targets of the other half: no target is unlearned with a potential it helped to select.

\addcontentsline{toc}{subsection}{The three steps}
\mypar{The three steps}
One evaluation of the meta-objective is a $24$-step run on the $24$ targets of a half, about $21$ minutes on one RTX~4090. In steps one and two, which used the meta-objective without its fidelity term, the measurements are read from the last two in-run checks. In step three the model after $24$ steps generates $16$ fresh images per target, with generator seeds shared by all candidates, and all terms of the meta-objective are computed from them; this adds about $20$ minutes per candidate.

The main paper describes the scheme of a step; the three steps implemented it as follows. \emph{Step one} moves each band gate by $\pm 1.5$ and the exponent, in one direction only, by $+0.3$ around $\psi = 0$ ($16$ runs). Then it evaluates six candidates along the descent direction, with the largest band gate moved by $3$, $6$ or $9$ and with or without the exponent step, repeats some of them with a second seed, and runs two hand-set low-pass potentials alongside as references. \emph{Step two} fits a quadratic model to all evaluations of the half and evaluates its minimizer and the midpoint between it and the current potential, partly with a second seed. \emph{Step three} moves every parameter by the same amounts in both directions around the current potential ($17$ runs). Then it re-runs the three best candidates and a candidate that combines the best-scoring moves, each with two further seeds. Finally, it continues along the exponent, the one direction that stood out.

\addcontentsline{toc}{subsection}{Variation between runs}
\mypar{Variation between runs}
Two runs with the same potential and the same seed differ by up to $0.08$ in the meta-objective, and the standard deviation over seeds is between $0.02$ and $0.09$ (Table~\ref{tab:step3}). Moving a single band gate changes the meta-objective by an amount of the same size, which is why step three compares means over three seeds and why we did not take further steps on the band gates. The exponent is the exception: lowering it by $0.3$ lowers the meta-objective by $0.11$ and $0.10$ on the two halves, and lowering it further raises the similarity to the memorized image above the floor again.

\begin{table}[t]
\centering
\footnotesize
\setlength{\tabcolsep}{2.6pt}
\begin{tabular}{llcccc}
\toprule
 & Candidate & $\mathcal{L}_{\mathrm{meta}}$ & SSCD & IQA & CLIP-I \\
\midrule
Half A & after step two & $0.510 \pm 0.029$ & $0.060$ & $0.743$ & $0.634$ \\
 & $g_0 - 1.5$ & $0.475 \pm 0.058$ & $0.083$ & $0.783$ & $0.651$ \\
 & $g_3 - 1.5$ & $0.489 \pm 0.030$ & $0.090$ & $0.755$ & $0.662$ \\
 & combined & $0.464 \pm 0.026$ & $0.087$ & $0.778$ & $0.663$ \\
 & $p - 0.3$ (kept) & $0.397 \pm 0.017$ & $0.090$ & $0.798$ & $0.679$ \\
\midrule
Half B & after step two & $0.490 \pm 0.085$ & $0.065$ & $0.808$ & $0.598$ \\
 & $g_4 + 1.5$ & $0.432 \pm 0.039$ & $0.066$ & $0.802$ & $0.628$ \\
 & $g_6 + 1.5$ & $0.406 \pm 0.032$ & $0.072$ & $0.832$ & $0.628$ \\
 & combined & $0.398 \pm 0.027$ & $0.085$ & $0.843$ & $0.650$ \\
 & $p - 0.3$ (kept) & $0.388 \pm 0.051$ & $0.079$ & $0.847$ & $0.642$ \\
 & $p - 0.6$ & $0.428 \pm 0.043$ & $0.118$ & $0.869$ & $0.672$ \\
 & $p - 0.9$ & $0.467 \pm 0.084$ & $0.140$ & $0.869$ & $0.677$ \\
\bottomrule
\end{tabular}
\caption{Step three of the search: candidates that were run with several seeds, on the half they are found on (mean $\pm$ standard deviation over three seeds; two seeds for the last two rows). The metrics are those of $16$ fresh generations per target (IQA: CLIP-IQA).}
\label{tab:step3}
\end{table}

\mypar{The first two steps}
Under the meta-objective without the fidelity term, read from the in-run checks, the squared-$L^2$ loss, the hand-set low-pass, and the potentials after steps one and two score $0.411$, $0.317$, $0.274$, $0.266$ on half A and $0.428$, $0.257$, $0.257$, $0.237$ on half B. Table~\ref{tab:learning} gives the same potentials under the full meta-objective.

\begin{table}[t]
\centering
\small
\begin{tabular}{lcc}
\toprule
Potential & Half A & Half B \\
\midrule
Squared-$L^2$ (start) & $0.541$ & $0.584$ \\
Hand-set low-pass & $0.475$ & $0.477$ \\
\midrule
BEND, one step & $0.515$ & $0.412$ \\
BEND, two steps & $0.510 \pm 0.029$ & $0.490 \pm 0.085$ \\
BEND, three steps & $0.397 \pm 0.017$ & $0.388 \pm 0.051$ \\
\bottomrule
\end{tabular}
\caption{The full meta-objective (lower is better) on the half of the targets a potential is found on. $\pm$: standard deviation over three seeds; other entries are single runs, for which a repetition can differ by up to $0.08$.}
\label{tab:learning}
\end{table}

\addcontentsline{toc}{subsection}{Sensitivity to the weights of the meta-objective}
\mypar{Sensitivity to the weights of the meta-objective}
The weights $\mu$, $\nu$, $\rho$ and the floors $\tau_c$ of the meta-objective were set by hand. To see how much the search depends on them, we re-scored every logged evaluation of the three steps, on both halves, under $81$ weightings: $\mu \in \{2, 4, 8\}$, $\nu \in \{1, 2, 4\}$ (in step three only, as the first two steps had no fidelity term), $\rho \in \{0, 1, 2\}$, and three choices of the floors (those of the main paper, half the mean similarity of the retain images to the memorized image, and $0.1$ for every target). A decision counts as unchanged when the candidate the search took is still the best of the candidates it evaluated; candidates that were never run cannot be judged. Table~\ref{tab:weights} gives the result. The direction of the search does not depend on the weights: under every weighting, the finite-difference gradient of step one has a cosine of at least $0.98$ to the original one. The weights move the choice along the trade-off between forgetting and fidelity, as they should. In step three, less weight on forgetting ($\mu = 2$) or more on fidelity ($\nu = 4$) mostly selects a larger change of the exponent, which forgets less (SSCD $0.12$ to $0.14$, against $0.08$ to $0.09$ for the selected potential) and keeps more fidelity; on half B, more weight on forgetting ($\mu = 8$) selects the move of the finest band gate, which is within $0.015$ of the selected potential in SSCD, CLIP-IQA, and CLIP-I. The weight $\rho$ and the floors alone never change the choice.

\begin{table}[t]
\centering
\small
\begin{tabular}{lcc}
\toprule
Decision of the search & Half A & Half B \\
\midrule
Step one, direction (cosine) & $\geq 0.98$ & $\geq 0.98$ \\
Step one, point on the line & $54$ & $48$ \\
Step two & $81$ & $78$ \\
Step three, in the best three & $81$ & $72$ \\
Step three, selected move & $69$ & $40$ \\
\quad of these with $\mu = 4$ & $27$ of $27$ & $19$ of $27$ \\
\bottomrule
\end{tabular}
\caption{Sensitivity of the search to the weights of the meta-objective: number of the $81$ weightings under which a decision of the search is unchanged, from a re-scoring of the logged evaluations (no new runs). For the direction of step one, the smallest cosine between the finite-difference gradients under the new and the original weights.}
\label{tab:weights}
\end{table}

\section{Additional results}
\label{app:results}

\begin{table*}[t]
\centering
\small
\setlength{\tabcolsep}{3.5pt}
\begin{tabular}{llrrrrrrr}
\toprule
 & & \multicolumn{2}{c}{Memorized prompt} & \multicolumn{2}{c}{Modified prompt} & \multicolumn{1}{c}{Unrelated} & \multicolumn{2}{c}{Prompt fidelity} \\
\cmidrule(lr){3-4} \cmidrule(lr){5-6} \cmidrule(lr){7-7} \cmidrule(lr){8-9}
Target & Potential & \multicolumn{1}{c}{CLIP-IQA $\uparrow$} & \multicolumn{1}{c}{SSCD $\downarrow$} & \multicolumn{1}{c}{CLIP-IQA $\uparrow$} & \multicolumn{1}{c}{SSCD $\downarrow$} & \multicolumn{1}{c}{CLIP-IQA $\uparrow$} & \multicolumn{1}{c}{CLIP-T $\uparrow$} & \multicolumn{1}{c}{CLIP-I $\uparrow$} \\
\midrule
ReTrack & squared-$L^2$, $24$ steps & $0.636 \pm 0.044$ & $0.094 \pm 0.003$ & $0.746$ & $0.081$ & $0.896$ & $0.248$ & $0.658$ \\
 & squared-$L^2$, stop rule & $0.637 \pm 0.037$ & $0.096 \pm 0.008$ & $0.741$ & $0.080$ & $0.897$ & $0.239$ & $0.646$ \\
 & hand-set & $0.771 \pm 0.032$ & $0.172 \pm 0.041$ & $0.843$ & $0.123$ & $0.921$ & $0.251$ & $0.661$ \\
\midrule
Retain mean & squared-$L^2$ & $0.682 \pm 0.026$ & $0.068 \pm 0.003$ & $0.746$ & $0.072$ & $0.901$ & $0.247$ & $0.653$ \\
 & hand-set & $0.791 \pm 0.013$ & $0.082 \pm 0.004$ & $0.827$ & $0.089$ & $0.918$ & $0.233$ & $0.631$ \\
 & searched, one step & $0.799 \pm 0.018$ & $0.073 \pm 0.003$ & $0.825$ & $0.089$ & $0.922$ & $0.237$ & $0.624$ \\
 & searched, two steps & $0.803 \pm 0.015$ & $0.063 \pm 0.005$ & $0.835$ & $0.082$ & $0.922$ & $0.231$ & $0.618$ \\
 & searched, three steps & $0.811 \pm 0.002$ & $0.084 \pm 0.004$ & $0.844$ & $0.098$ & $0.922$ & $0.246$ & $0.651$ \\
\midrule
Tilted ($\kappa = 0.10$) & hand-set & $0.781$ & $0.069$ & $0.822$ & $0.085$ & $0.925$ & $0.230$ & $0.623$ \\
 & searched, one step & $0.782$ & $0.065$ & $0.820$ & $0.086$ & $0.918$ & $0.238$ & $0.623$ \\
\bottomrule
\end{tabular}
\caption{All jointly trained systems at $N = 48$, by redirection target and potential, in the format of Table~1 of the main paper. $\pm$: standard deviation over three seeds (two for the stop rule, whose fidelity is from one seed); other rows are one seed. \emph{Searched, three steps} is BEND.}
\label{tab:full}
\end{table*}

\begin{figure*}[t]
\centering
\includegraphics[width=0.92\textwidth]{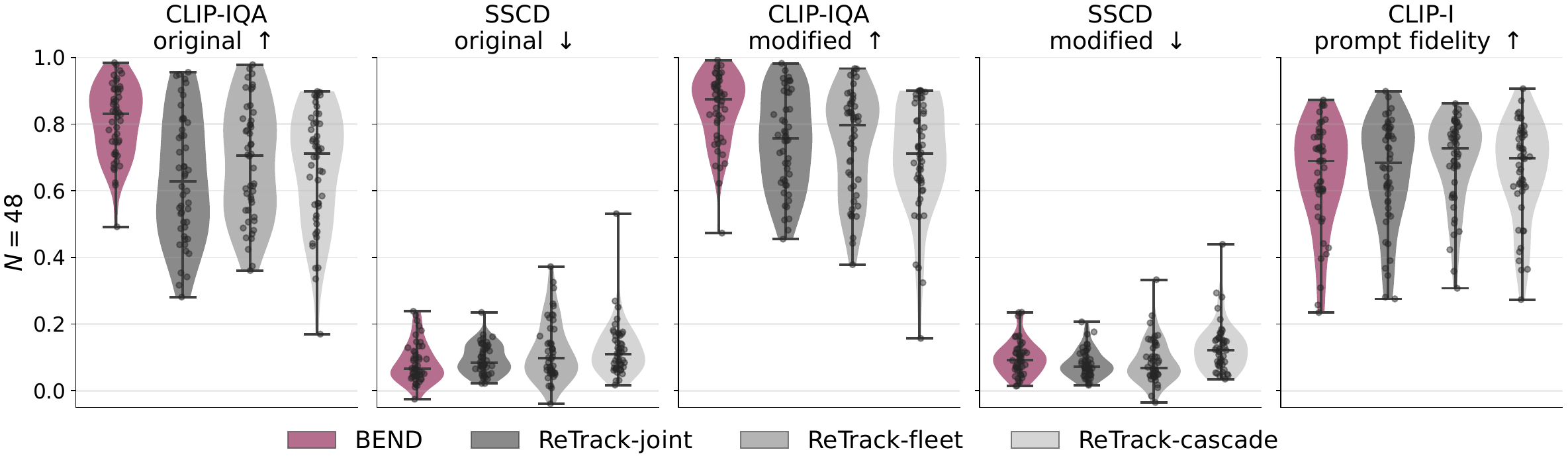}
\caption{Per-target distributions at $N = 48$: CLIP-IQA and SSCD on the original and modified prompts, and prompt fidelity (CLIP-I); one shared $y$ scale. Violins: distribution; points: targets; bars: median and range.}
\label{fig:sd-violins48}
\end{figure*}

\begin{figure*}[t]
\centering
\includegraphics[width=0.85\textwidth]{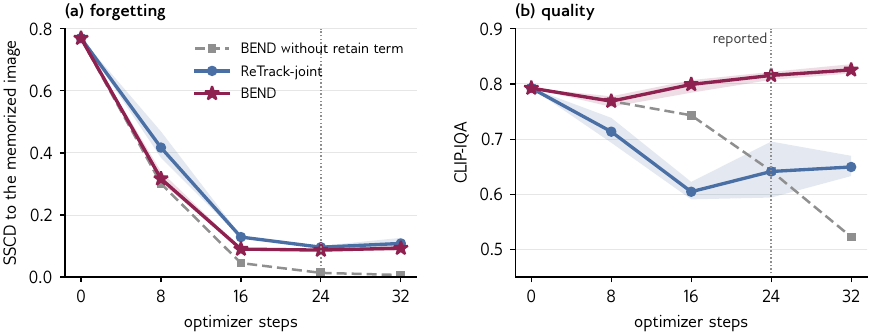}
\caption{Forgetting and quality over the unlearning run at $N = 48$, from the checks during unlearning (eight generations per target, mean over the targets; bands: range over three seeds). Step $0$ is the pretrained model. The dotted line marks the $24$ steps after which all results are reported.}
\label{fig:sd-curves}
\end{figure*}

\begin{figure*}[tp]
\centering
\includegraphics[width=\textwidth]{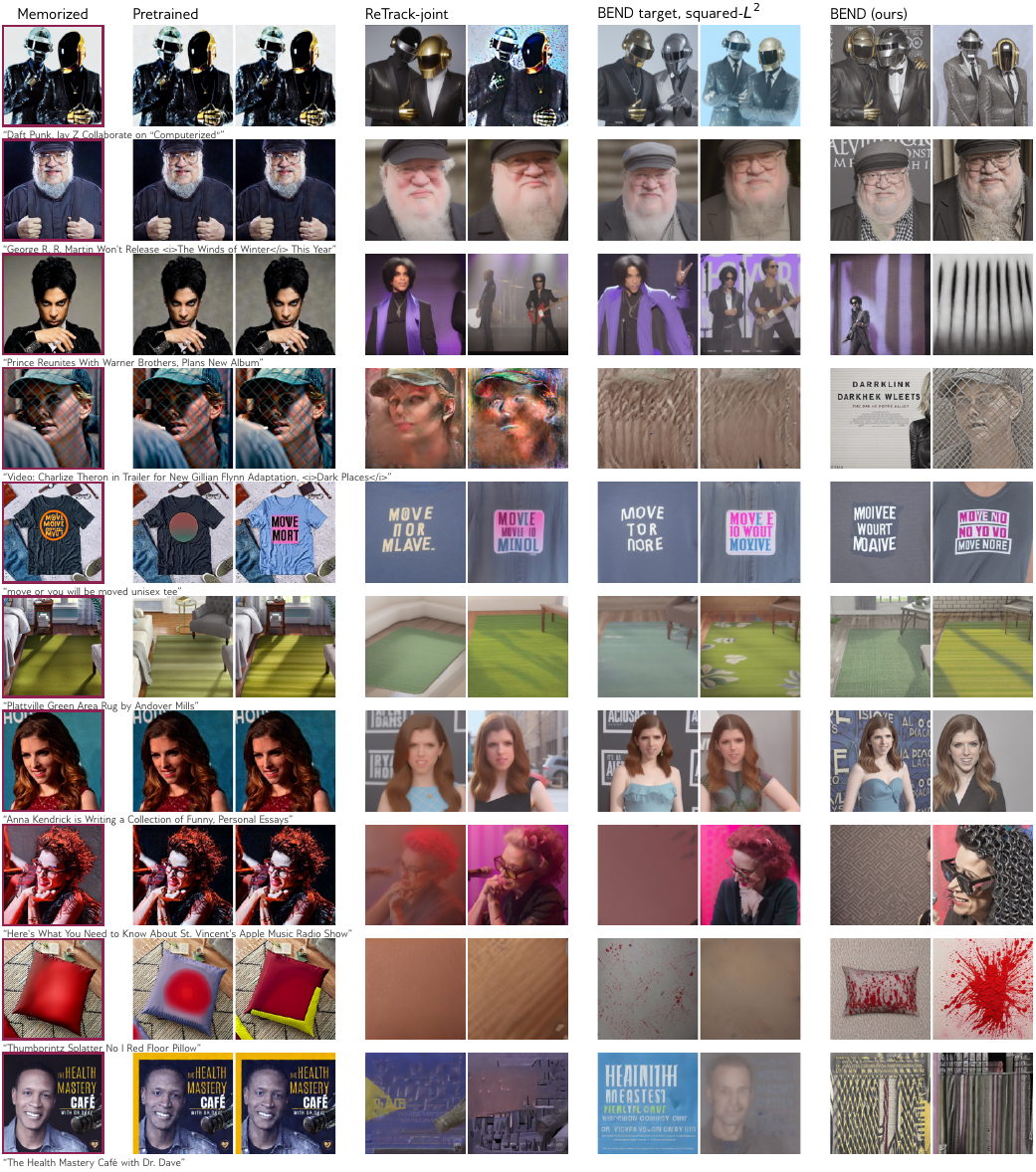}
\caption{Stable Diffusion generations for ten forget targets (two generations per system, benchmark sampler). Each row is one target: the memorized training image with its prompt, then two generations from each system. The targets were chosen without looking at images, at evenly spaced positions in the ranking of the targets by the CLIP-IQA difference between BEND and ReTrack-joint, from the target where BEND gains least (top) to where it gains most (bottom). Pretrained reproduces the memorized image; ReTrack-joint and the squared-$L^2$ loss with the BEND target often forget with blurred or degraded images; BEND gives sharp images; where it fails, the image does not show the content of the prompt, the loss of fidelity discussed in the main paper.}
\label{fig:sd-samples}
\end{figure*}

This appendix reports the paired comparisons of the main paper with their intervals (Table~\ref{tab:paired}), all jointly trained systems (Table~\ref{tab:full}), ablations of the target and of the retain term (Table~\ref{tab:ablation}, Fig.~\ref{fig:sd-curves}), per-target statistics (Table~\ref{tab:pertarget}), the dependence on the length of the run (Table~\ref{tab:length}) and on the sampler (Table~\ref{tab:sampler}), generations for a set of targets (Fig.~\ref{fig:sd-samples}), measures that the search never saw (Tables~\ref{tab:heldout} and~\ref{tab:unrelated}), unrelated COCO captions (Table~\ref{tab:coco}), the mask of \citet{S-park2025frequency} (Table~\ref{tab:pp}), and relearning (Table~\ref{tab:relearn}). Fig.~\ref{fig:sd-violins48} shows the distribution over the $48$ targets behind the means of the main paper.

\begin{table}[t]
\centering
\small
\setlength{\tabcolsep}{3pt}
\begin{tabular}{lrc}
\toprule
Metric & Difference & $95\%$ interval \\
\midrule
\multicolumn{3}{l}{\emph{BEND $-$ ReTrack-joint}} \\
CLIP-IQA & $+0.174$ & $[+0.135, +0.215]$ \\
SSCD & $-0.010$ & $[-0.022, +0.002]$ \\
CLIP-T & $-0.002$ & $[-0.012, +0.009]$ \\
CLIP-I & $-0.008$ & $[-0.022, +0.007]$ \\
\multicolumn{3}{l}{\emph{BEND $-$ ReTrack-fleet}} \\
SSCD & $-0.036$ & $[-0.064, -0.011]$ \\
CLIP-T & $-0.013$ & $[-0.025, -0.000]$ \\
CLIP-I & $-0.039$ & $[-0.060, -0.018]$ \\
\multicolumn{3}{l}{\emph{BEND $-$ ReTrack-cascade}} \\
SSCD & $-0.026$ & $[-0.049, -0.005]$ \\
\midrule
\multicolumn{3}{l}{\emph{BEND $-$ hand-set $\Phi$}} \\
CLIP-IQA & $+0.019$ & $[+0.001, +0.036]$ \\
CLIP-T & $+0.013$ & $[+0.007, +0.019]$ \\
CLIP-I & $+0.020$ & $[+0.011, +0.029]$ \\
\multicolumn{3}{l}{\emph{No fidelity term $-$ ReTrack-joint}} \\
CLIP-T & $-0.017$ & $[-0.029, -0.006]$ \\
CLIP-I & $-0.040$ & $[-0.058, -0.022]$ \\
\bottomrule
\end{tabular}
\caption{Paired comparisons at $N = 48$ quoted in the main paper: mean over the $48$ targets of the per-target difference of seed-averaged values, and its bootstrap $95\%$ confidence interval. \emph{Hand-set $\Phi$}: with the BEND target; \emph{no fidelity term}: BEND with the potential found without the fidelity term. The fidelity of ReTrack-fleet is from one of its seeds.}
\label{tab:paired}
\end{table}

\mypar{The tilted target}
Tilting the target away from the memorized image (Appendix~\ref{app:redirect}) forgets more at a small loss of quality: under the hand-set potential SSCD falls from $0.082$ to $0.069$ and CLIP-IQA from $0.791$ to $0.781$ (Table~\ref{tab:full}).

\addcontentsline{toc}{subsection}{Ablations}
\mypar{Ablations}
Table~\ref{tab:ablation} changes one component of BEND at a time, with BEND's potential kept, and Fig.~\ref{fig:sd-curves} follows the run over its steps. The first two rows concern the number of retain images behind the target. \citet{S-shi2026retrack} vary the number $k$ of neighbors and find that their unlearning metrics do not change and quality improves only slightly, which the collapse of the weights explains: whatever $k$ is, the weights select one image (Appendix~\ref{app:redirect}). With the mean in place of the weights, the number of images matters. Under ReTrack's weights over the $10$ nearest retain images SSCD is $0.138$, with the mean of the same $10$ images it is $0.091$, and with the mean of all retain images, $66$ on average, $0.081$. Quality is the same for the three targets (CLIP-IQA $0.79$ to $0.81$), i.e., it is set by the potential, as for the ablations in the main paper. The last row removes the retain term ($\lambda = 1$). The model then forgets almost everything about the targets (SSCD $0.019$), but it loses quality (CLIP-IQA $0.633$), fidelity (CLIP-T $0.157$), and, in contrast to every other system, also quality on unrelated prompts ($0.807$ against $0.922$). Fig.~\ref{fig:sd-curves} shows that this loss grows with every step. With the retain term, BEND reaches its level of forgetting after $16$ steps and its quality does not fall below the $0.77$ of the first check, whereas ReTrack-joint loses quality during the first $16$ steps and recovers only part of it.

\begin{table*}[t]
\centering
\small
\setlength{\tabcolsep}{3.5pt}
\begin{tabular}{lrrrrrrr}
\toprule
 & \multicolumn{2}{c}{Memorized prompt} & \multicolumn{2}{c}{Modified prompt} & \multicolumn{1}{c}{Unrelated} & \multicolumn{2}{c}{Prompt fidelity} \\
\cmidrule(lr){2-3} \cmidrule(lr){4-5} \cmidrule(lr){6-6} \cmidrule(lr){7-8}
System & \multicolumn{1}{c}{CLIP-IQA $\uparrow$} & \multicolumn{1}{c}{SSCD $\downarrow$} & \multicolumn{1}{c}{CLIP-IQA $\uparrow$} & \multicolumn{1}{c}{SSCD $\downarrow$} & \multicolumn{1}{c}{CLIP-IQA $\uparrow$} & \multicolumn{1}{c}{CLIP-T $\uparrow$} & \multicolumn{1}{c}{CLIP-I $\uparrow$} \\
\midrule
BEND & $0.807$ & $0.081$ & $0.842$ & $0.094$ & $0.922$ & $0.245$ & $0.650$ \\
\midrule
Target: mean of the $10$ nearest retain images & $0.794$ & $0.091$ & $0.847$ & $0.103$ & $0.917$ & $0.240$ & $0.639$ \\
Target: ReTrack's weights over the $10$ nearest & $0.810$ & $0.138$ & $0.847$ & $0.108$ & $0.923$ & $0.249$ & $0.668$ \\
\midrule
Without the retain term ($\lambda = 1$) & $0.633$ & $0.019$ & $0.614$ & $0.019$ & $0.807$ & $0.157$ & $0.508$ \\
\bottomrule
\end{tabular}
\caption{Ablations of BEND at $N = 48$: the redirection target and the retain term, with BEND's potential and the training seed of the first row (one seed each). One target keeps a replicated generation under each of the two changed targets, none in the other rows.}
\label{tab:ablation}
\end{table*}

\addcontentsline{toc}{subsection}{Per-target statistics}
\mypar{Per-target statistics}
A favorable mean does not guarantee that every deletion request is honored, so Table~\ref{tab:pertarget} looks at the targets one by one. A model that had never memorized an image would still produce images with some similarity to it, namely that of the retain images of its prompt; we count the targets whose SSCD remains above this level. For ReTrack these are $5$ to $13$ targets, for BEND $2$ to $3$. The worst single target is similar across the joint systems (SSCD between $0.21$ and $0.30$). For ReTrack-joint it is always the same target, whose retain images are themselves close to the memorized image (similarity $0.33$), so that it is in fact below its own retain level. With the BEND target it is, in half of the runs, a second target that stays above its retain level ($0.13$) and is not fully forgotten.

\begin{table}[t]
\centering
\small
\setlength{\tabcolsep}{3pt}
\begin{tabular}{lcccc}
\toprule
 & Worst & Above & Repli- & CLIP-IQA \\
System & SSCD & retain level & cated & $< 0.5$ \\
\midrule
ReTrack-fleet & $0.37$--$0.38$ & $10$--$11$ & $0$--$1$ & $7$--$9$ \\
ReTrack-cascade & $0.25$--$0.53$ & $11$--$13$ & $0$--$1$ & $6$--$9$ \\
ReTrack-joint & $0.21$--$0.27$ & $5$--$7$ & $0$ & $8$--$18$ \\
Hand-set $\Phi$ & $0.21$--$0.29$ & $1$--$3$ & $0$ & $0$--$1$ \\
BEND, one step & $0.24$--$0.28$ & $2$--$3$ & $0$--$1$ & $0$--$1$ \\
BEND, two steps & $0.21$--$0.28$ & $1$--$2$ & $0$ & $0$--$1$ \\
BEND, three steps & $0.25$--$0.30$ & $2$--$3$ & $0$--$1$ & $0$--$1$ \\
\bottomrule
\end{tabular}
\caption{Per-target statistics at $N = 48$ (ranges over three seeds where available). \emph{Worst SSCD}: highest per-target mean. \emph{Above retain level}: targets whose SSCD exceeds the mean similarity of their own retain images to the memorized image. \emph{Replicated}: targets with at least one of $64$ generations above SSCD $0.5$. \emph{CLIP-IQA $< 0.5$}: targets with degraded generations. \emph{Hand-set $\Phi$}: with the BEND target.}
\label{tab:pertarget}
\end{table}

\mypar{Diversity}
The mean SSCD similarity among the $64$ generations of one prompt is $0.16$ to $0.20$ for BEND and for the hand-set potential, and $0.22$ to $0.25$ for ReTrack-joint. For reference, it is $0.18$ among the retain images of a prompt and $0.81$ among its memorized generations.

\addcontentsline{toc}{subsection}{Prompt fidelity}
\mypar{Prompt fidelity}
Among the single models, BEND and ReTrack-joint are level in fidelity: the paired differences, $-0.002$ in CLIP-T and $-0.008$ in CLIP-I, are not significant (Table~\ref{tab:paired}). The fleet, which spends one model on every target, keeps the most; it is above BEND by $0.013$ in CLIP-T, at the edge of significance, and significantly by $0.039$ in CLIP-I. The level of BEND is the effect of the fidelity term of the meta-objective: without it, BEND is significantly below ReTrack-joint, by $0.017$ in CLIP-T and by $0.040$ in CLIP-I. The failures of the two methods still look different (Fig.~\ref{fig:sd-samples}). Where ReTrack-joint fails it produces a blurred image, and where BEND fails it produces a sharp image of the wrong content, which CLIP-IQA rewards.

\mypar{Length of the run}
Table~\ref{tab:length} evaluates the stored models after $16$, $24$, and $32$ steps. With the BEND target the two sides of unlearning settle at different times: forgetting and CLIP-IQA are in place after $16$ steps, while prompt fidelity is lowest there and recovers as the run continues. ReTrack-joint reaches its fidelity after $24$ steps and does not improve further, and its SSCD rises again.

\begin{table}[t]
\centering
\small
\setlength{\tabcolsep}{3pt}
\begin{tabular}{lrcccc}
\toprule
System & Steps & CLIP-T & CLIP-I & IQA & SSCD \\
\midrule
ReTrack-joint & $16$ & $0.227$ & $0.621$ & $0.631$ & $0.127$ \\
 & $24$ & $0.248$ & $0.658$ & $0.670$ & $0.092$ \\
 & $32$ & $0.250$ & $0.659$ & $0.651$ & $0.113$ \\
\midrule
Hand-set $\Phi$ & $24$ & $0.233$ & $0.631$ & $0.776$ & $0.080$ \\
 & $32$ & $0.246$ & $0.651$ & $0.788$ & $0.086$ \\
\midrule
BEND, one step & $16$ & $0.219$ & $0.594$ & $0.771$ & $0.071$ \\
 & $24$ & $0.237$ & $0.624$ & $0.796$ & $0.074$ \\
 & $32$ & $0.245$ & $0.638$ & $0.786$ & $0.079$ \\
\midrule
BEND, two steps & $24$ & $0.231$ & $0.618$ & $0.796$ & $0.065$ \\
 & $32$ & $0.236$ & $0.625$ & $0.796$ & $0.069$ \\
\midrule
BEND, three steps & $24$ & $0.246$ & $0.651$ & $0.805$ & $0.088$ \\
 & $32$ & $0.253$ & $0.664$ & $0.821$ & $0.092$ \\
\bottomrule
\end{tabular}
\caption{Dependence on the length of the unlearning run, measured on the $16$ fidelity generations per prompt (seed means; rows with $16$ steps are one seed). \emph{Hand-set $\Phi$}: with the BEND target. IQA: CLIP-IQA.}
\label{tab:length}
\end{table}

\addcontentsline{toc}{subsection}{A second sampler}
\mypar{A second sampler}
Table~\ref{tab:sampler} repeats the fidelity evaluation with the default sampler of the public Stable Diffusion pipeline (PNDM, $50$ steps). For the memorized prompts this sampler gives saturated images with strong artifacts already for the pretrained model, and it reproduces the memorized image less often (SSCD $0.42$ instead of $0.76$), so neither memorization nor its removal is measured cleanly with it. We report it because it is what a user of the public pipeline gets. The outcome is mixed: the differences in CLIP-IQA are smaller, and only with the BEND target does a low-pass potential stay ahead of the systems with the squared-$L^2$ loss; with ReTrack's target the hand-set potential falls to their level. The systems with a low-pass potential also lose more fidelity than under the benchmark sampler.

\begin{table}[t]
\centering
\footnotesize
\setlength{\tabcolsep}{1.5pt}
\begin{tabular}{lcccccccc}
\toprule
 & \multicolumn{4}{c}{Benchmark sampler} & \multicolumn{4}{c}{Pipeline default} \\
\cmidrule(lr){2-5} \cmidrule(lr){6-9}
System & IQA & SSCD & T & I & IQA & SSCD & T & I \\
\midrule
Pretrained & $.794$ & $.763$ & $.287$ & $.723$ & $.546$ & $.415$ & $.236$ & $.626$ \\
ReTrack-cascade & $.700$ & $.111$ & $.241$ & $.651$ & $.654$ & $.100$ & $.226$ & $.628$ \\
ReTrack-joint & $.670$ & $.092$ & $.248$ & $.658$ & $.657$ & $.089$ & $.233$ & $.637$ \\
ReTrack target, hand-set & $.778$ & $.171$ & $.251$ & $.661$ & $.652$ & $.119$ & $.218$ & $.610$ \\
BEND target, $L^2$ & $.687$ & $.070$ & $.247$ & $.653$ & $.657$ & $.068$ & $.231$ & $.630$ \\
BEND target, hand-set & $.776$ & $.080$ & $.233$ & $.631$ & $.727$ & $.076$ & $.223$ & $.611$ \\
BEND, one step & $.796$ & $.074$ & $.237$ & $.624$ & $.699$ & $.068$ & $.216$ & $.592$ \\
\bottomrule
\end{tabular}
\caption{The same models under two samplers ($16$ generations per prompt; IQA: CLIP-IQA; T: CLIP-T; I: CLIP-I). Values are means over the seeds evaluated with $16$ generations, i.e., those of the prompt-fidelity columns of Table~1 of the main paper; the pretrained model is one run.}
\label{tab:sampler}
\end{table}

\addcontentsline{toc}{subsection}{Runtime}
\mypar{Runtime}
A joint run on all $48$ targets ($32$ steps; the reported model is the one after $24$) takes between $39$ and $49$ minutes on one RTX~4090, including the in-run checks (six runs measured). The search for one potential took about $11$ GPU-hours for steps one and two ($30$ to $32$ runs of about $21$ minutes, including the repeats and the reference potentials) and about $24$ GPU-hours for step three (about $36$ runs of $40$ minutes, which include the scoring of fresh generations). For comparison, one model of ReTrack-fleet takes about $7$ minutes, i.e., $5.7$ GPU-hours for the $48$ targets. This includes a check on $16$ generations after every step, which we ran for monitoring and which ReTrack does not need.

\addcontentsline{toc}{subsection}{Measures outside the search}
\mypar{Measures outside the search}
The search scores a potential with SSCD, CLIP-IQA, and CLIP-I, and the main paper evaluates with the same SSCD and CLIP-IQA; it names the transfer to held-out measures as a limitation. As a partial check, we scored the images behind Table~1 of the main paper, regenerated with the same seeds (they reproduce the published SSCD and CLIP-IQA exactly for $21$ of the $22$ models, and within $0.013$ for the remaining one), with two further measures of forgetting, the DINOv2 \citep{S-oquab2024dinov2} and CLIP image similarity to the memorized image, and with two no-reference quality measures, MUSIQ \citep{S-ke2021musiq} and TOPIQ \citep{S-chen2024topiq}. Table~\ref{tab:heldout} gives the results. On the memorized prompt, by all four measures BEND forgets at least as much as ReTrack-joint and ReTrack-cascade and keeps a higher quality. Against ReTrack-joint the paired differences are $-0.060$ $[-0.085, -0.035]$ in DINOv2 and $-0.014$ $[-0.026, -0.002]$ in CLIP similarity, both significant, and $+8.7$ $[+6.5, +11.1]$ in MUSIQ and $+0.074$ $[+0.056, +0.094]$ in TOPIQ. Against ReTrack-cascade BEND is lower by $0.089$ $[0.037, 0.141]$ in DINOv2, level in CLIP similarity ($-0.015$ $[-0.043, +0.013]$), and higher by $6.1$ $[3.6, 8.6]$ in MUSIQ and $0.123$ $[0.092, 0.153]$ in TOPIQ. With the BEND target, BEND is level with the squared-$L^2$ loss in forgetting and higher in MUSIQ by $4.5$ $[2.6, 6.8]$. Against the hand-set potential with the same target, BEND is level in forgetting (DINOv2 $-0.006$ $[-0.020, +0.008]$) and slightly lower in these two quality measures (MUSIQ $-0.9$ $[-1.6, -0.3]$, TOPIQ $-0.013$ $[-0.018, -0.007]$). On the twelve unrelated prompts (Table~\ref{tab:unrelated}), BEND and ReTrack-joint move the generations equally far from those of the pretrained model with the same noise (LPIPS \citep{S-zhang2018lpips} $0.262$ and $0.257$), and ReTrack-cascade about twice as far ($0.498$); only ReTrack-cascade loses quality on them by these measures.

\begin{table*}[t]
\centering
\small
\setlength{\tabcolsep}{4pt}
\begin{tabular}{lrrrrrrrr}
\toprule
 & \multicolumn{4}{c}{Memorized prompt} & \multicolumn{4}{c}{Modified prompt} \\
\cmidrule(lr){2-5} \cmidrule(lr){6-9}
System & DINOv2 $\downarrow$ & CLIP $\downarrow$ & MUSIQ $\uparrow$ & TOPIQ $\uparrow$ & DINOv2 $\downarrow$ & CLIP $\downarrow$ & MUSIQ $\uparrow$ & TOPIQ $\uparrow$ \\
\midrule
Pretrained & $0.898$ & $0.938$ & $63.9$ & $0.457$ & $0.742$ & $0.857$ & $66.6$ & $0.511$ \\
ReTrack-cascade & $0.422$ & $0.680$ & $62.6$ & $0.553$ & $0.436$ & $0.692$ & $65.4$ & $0.554$ \\
ReTrack-joint & $0.394$ & $0.679$ & $60.0$ & $0.601$ & $0.392$ & $0.689$ & $67.9$ & $0.658$ \\
BEND (ours) & $0.333$ & $0.664$ & $68.7$ & $0.676$ & $0.374$ & $0.686$ & $72.4$ & $0.708$ \\
\midrule
ReTrack target, hand-set $\Phi$ & $0.475$ & $0.718$ & $68.5$ & $0.656$ & $0.422$ & $0.709$ & $73.1$ & $0.720$ \\
BEND target, squared-$L^2$ & $0.333$ & $0.660$ & $64.2$ & $0.630$ & $0.361$ & $0.677$ & $68.4$ & $0.658$ \\
BEND target, hand-set $\Phi$ & $0.340$ & $0.662$ & $69.6$ & $0.688$ & $0.372$ & $0.679$ & $72.4$ & $0.711$ \\
BEND, no fidelity term & $0.285$ & $0.632$ & $68.6$ & $0.687$ & $0.335$ & $0.659$ & $71.9$ & $0.712$ \\
\bottomrule
\end{tabular}
\caption{Measures that the search never saw, on the images behind Table~1 of the main paper ($64$ generations per prompt; means over the $48$ targets and over the seeds of Table~1; the pretrained model is one run). DINOv2 and CLIP: cosine similarity of the DINOv2-base and CLIP ViT-L/14 image embeddings to the memorized image (lower is more forgotten; this CLIP similarity is not the fidelity measure CLIP-I). MUSIQ and TOPIQ: no-reference quality (KonIQ models).}
\label{tab:heldout}
\end{table*}

\begin{table}[t]
\centering
\small
\setlength{\tabcolsep}{4pt}
\begin{tabular}{lrrr}
\toprule
System & LPIPS $\downarrow$ & MUSIQ $\uparrow$ & TOPIQ $\uparrow$ \\
\midrule
Pretrained & -- & $71.9$ & $0.641$ \\
ReTrack-cascade & $0.498$ & $69.8$ & $0.609$ \\
ReTrack-joint & $0.257$ & $73.2$ & $0.743$ \\
BEND (ours) & $0.262$ & $74.1$ & $0.746$ \\
\midrule
ReTrack target, hand-set $\Phi$ & $0.229$ & $73.9$ & $0.734$ \\
BEND target, squared-$L^2$ & $0.285$ & $73.4$ & $0.746$ \\
BEND target, hand-set $\Phi$ & $0.268$ & $74.1$ & $0.754$ \\
BEND, no fidelity term & $0.330$ & $74.4$ & $0.758$ \\
\bottomrule
\end{tabular}
\caption{The twelve unrelated prompts of the main paper, eight generations each (the first four are those of its unrelated evaluation; means over the seeds of Table~1). LPIPS: distance to the generation of the pretrained model with the same noise.}
\label{tab:unrelated}
\end{table}

\addcontentsline{toc}{subsection}{Unrelated prompts at scale}
\mypar{Unrelated prompts at scale}
Twelve unrelated prompts are few. We therefore generated one image for each of $1000$ captions of the COCO test split \citep{S-lin2014coco,S-karpathy2015deep} with each one-model system (three seeds) and with the pretrained model, with the same noise for all, and compared the images of each system with those of the pretrained model (Table~\ref{tab:coco}). KID \citep{S-binkowski2018demystifying} is zero within its noise for BEND and ReTrack-joint and positive for ReTrack-cascade. FID against the pretrained images of the same noise is $35.0$ for BEND, $32.9$ for ReTrack-joint, and $53.1$ for ReTrack-cascade; for scale, a second set of pretrained images with fresh noise has an FID of $48.6$ against the first. The similarity to the caption is unchanged except for a small loss of ReTrack-cascade. At this size, BEND and ReTrack-joint are indistinguishable from the pretrained model, while ReTrack-cascade drifts away from it.

\begin{table}[t]
\centering
\small
\begin{tabular}{lrrr}
\toprule
System & KID & FID & CLIP-T \\
\midrule
Pretrained, fresh noise & $-0.0003$ & $48.6$ & $0.263$ \\
ReTrack-cascade & $0.0019$ & $53.1$ & $0.261$ \\
ReTrack-joint & $-0.0001$ & $32.9$ & $0.265$ \\
BEND (ours) & $-0.0002$ & $35.0$ & $0.265$ \\
\bottomrule
\end{tabular}
\caption{$1000$ COCO captions, one generation each (pipeline sampler; means over three seeds). KID and FID against the images of the pretrained model for the same captions and noise; the first row uses fresh noise and shows the level between two sets of the same model. CLIP-T is the similarity to the COCO caption; for the pretrained model it is $0.264$.}
\label{tab:coco}
\end{table}

\addcontentsline{toc}{subsection}{The mask of Park and Park}
\mypar{The mask of Park and Park}
\citet{S-park2025frequency} restrict the forget term of loss-ascent methods to low spatial frequencies and to high noise levels. The hand-set ablation of the main paper takes the frequency part of this idea, a low-pass of the forget term at every noise level. We also ran their mask itself in the joint procedure, with the BEND target and with ReTrack's target (three seeds each, evaluated after $24$ steps like the hand-set ablations): the residual of the forget term passes a hard circular low-pass at $0.3$ of the Nyquist frequency, and the forget term acts only for $t \geq 750$. Their paper leaves three choices open, which we made as follows: the filter acts in the latent space of Stable Diffusion, where the loss is computed; the cutoff $r = 0.15$ is read in cycles per sample, i.e., $0.3$ of Nyquist; and the timestep window is a weight of $4$ inside the window and $0$ outside, which in expectation equals sampling $t$ in the window. Table~\ref{tab:pp} gives the result. In this procedure the mask barely forgets: SSCD stays at $0.439$ with the BEND target and $0.526$ with ReTrack's target, and $27$ to $41$ of the $48$ targets keep a replicated generation, depending on target and seed. The in-run checks show that forgetting stops after the first eight steps, presumably because the forget term never sees the lower noise levels, through which the memorized images return. The prompt fidelity of these models is high because they still largely produce the memorized content. Against both, BEND forgets more (paired SSCD $-0.355$ $[-0.407, -0.305]$ and $-0.442$ $[-0.490, -0.394]$) at higher quality (CLIP-IQA $+0.180$ $[0.123, 0.236]$ and $+0.133$ $[0.084, 0.182]$). Their mask was designed for methods that ascend the loss; this result concerns its use with the redirection targets of this paper.

\begin{table*}[t]
\centering
\small
\setlength{\tabcolsep}{3.5pt}
\begin{tabular}{lrrrrrrr}
\toprule
 & \multicolumn{2}{c}{Memorized prompt} & \multicolumn{2}{c}{Modified prompt} & \multicolumn{1}{c}{Unrelated} & \multicolumn{2}{c}{Prompt fidelity} \\
\cmidrule(lr){2-3} \cmidrule(lr){4-5} \cmidrule(lr){6-6} \cmidrule(lr){7-8}
System & \multicolumn{1}{c}{CLIP-IQA $\uparrow$} & \multicolumn{1}{c}{SSCD $\downarrow$} & \multicolumn{1}{c}{CLIP-IQA $\uparrow$} & \multicolumn{1}{c}{SSCD $\downarrow$} & \multicolumn{1}{c}{CLIP-IQA $\uparrow$} & \multicolumn{1}{c}{CLIP-T $\uparrow$} & \multicolumn{1}{c}{CLIP-I $\uparrow$} \\
\midrule
BEND (ours) & $0.811 \pm 0.002$ & $0.084 \pm 0.004$ & $0.844$ & $0.098$ & $0.922$ & $0.246$ & $0.651$ \\
Mask of~\cite{S-park2025frequency}, BEND target & $0.630 \pm 0.009$ & $0.439 \pm 0.010$ & $0.681$ & $0.313$ & $0.905$ & $0.271$ & $0.694$ \\
Mask of~\cite{S-park2025frequency}, ReTrack target & $0.678 \pm 0.013$ & $0.526 \pm 0.004$ & $0.706$ & $0.377$ & $0.908$ & $0.274$ & $0.702$ \\
\bottomrule
\end{tabular}
\caption{The mask of \citet{S-park2025frequency} in the joint procedure, in the format of Table~1 of the main paper (the BEND row is copied from there). $\pm$: standard deviation over three seeds. The comparisons in the main paper refer to the systems of its Table~1.}
\label{tab:pp}
\end{table*}

\addcontentsline{toc}{subsection}{Relearning}
\mypar{Relearning}
A model that has unlearned an image may relearn it from data close to it. We continued to fine-tune the unlearned models of BEND and ReTrack-joint (three seeds each) on the retain images of the $48$ targets with their prompts, with the standard diffusion loss and the optimizer of the unlearning run ($8$ samples per step), and scored them after $25$, $50$, and $100$ steps ($16$ generations per prompt). Table~\ref{tab:relearn} gives the result. Neither method is robust to this: in $100$ steps the mean SSCD to the memorized image roughly doubles, and up to $8$ targets per seed replicate again. Both relearn at the same rate; the paired differences are not significant at any step. The same fine-tuning also raises the CLIP-IQA of ReTrack-joint, which after $25$ steps is at the level of BEND in both SSCD and CLIP-IQA.

\begin{table}[t]
\centering
\small
\setlength{\tabcolsep}{3pt}
\begin{tabular}{rrcrrcr}
\toprule
 & \multicolumn{3}{c}{BEND} & \multicolumn{3}{c}{ReTrack-joint} \\
\cmidrule(lr){2-4} \cmidrule(lr){5-7}
Steps & SSCD & Repl. & IQA & SSCD & Repl. & IQA \\
\midrule
$0$ & $0.088$ & $0/0/0$ & $0.805$ & $0.093$ & $0/0/0$ & $0.646$ \\
$25$ & $0.107$ & $1/0/0$ & $0.757$ & $0.106$ & $0/3/1$ & $0.767$ \\
$50$ & $0.124$ & $4/0/0$ & $0.776$ & $0.124$ & $3/4/2$ & $0.778$ \\
$100$ & $0.166$ & $8/7/1$ & $0.784$ & $0.159$ & $6/8/5$ & $0.782$ \\
\bottomrule
\end{tabular}
\caption{Relearning from the retain images: SSCD to the memorized image and CLIP-IQA after continued fine-tuning. The scores use $16$ generations per prompt, so step $0$ differs slightly from Table~1 of the main paper, which uses $64$; means over three seeds. Repl.: targets with a generation above SSCD $0.5$, per seed.}
\label{tab:relearn}
\end{table}

\section{Unconditional benchmarks}
\label{app:ddpm}

This appendix gives the protocol and the complete results for the three unconditional benchmarks of \citet{S-shi2026retrack}, namely MNIST T-Shirt, CIFAR-10, and CelebA-HQ. We first describe the benchmarks, then how BEND is applied to a model without prompts and how its loss is searched for there, and finally the results. Note that the search has little to work with on these benchmarks: the forget term enters with a weight $\lambda$ of at most $0.02$, so quality is decided by the retain term.

\mypar{Benchmarks}
The three models are pixel-space DDPMs \citep{S-ho2020ddpm} without conditioning, and we keep the settings of \citet{S-shi2026retrack} for each of them. In MNIST T-Shirt the model was trained on MNIST with T-shirt images from Fashion-MNIST added at a ratio of $1\%$, and the T-shirt class is to be forgotten ($50$ optimizer steps, batch size $128$, learning rate $5 \cdot 10^{-5}$, $\lambda = 0.005$). The experiment is repeated with ten seeds. In CIFAR-10 ten training images and in CelebA-HQ ten identities are unlearned, each in a run of its own that starts from the pretrained model. A run has $60$ steps with batch size $128$, learning rate $2 \cdot 10^{-5}$, and $\lambda = 0.02$ on CIFAR-10, and $40$ steps with batch size $64$, learning rate $5 \cdot 10^{-6}$, and $\lambda = 0.005$ on CelebA-HQ. We ran ReTrack ourselves with these settings, so that both methods are measured with the same code and evaluation.

\mypar{Metrics}
Forgetting is measured by the negative log-likelihood (NLL) of the forgotten image under the model in bits per dimension (higher is better). On MNIST it is also measured by the frequency of the T-shirt among the generations, and on CIFAR-10 and CelebA-HQ by SSCD between the forgotten image and its reconstruction by the model after noising it to a quarter of the diffusion steps. Quality is measured by FID \citep{S-heusel2017fid} and, except on CelebA-HQ, by the Inception score (IS) \citep{S-salimans2016improved} of unconditional generations: $50{,}000$ generations with $50$ sampling steps on MNIST, and $10{,}000$ on CIFAR-10 ($1000$ steps) and CelebA-HQ ($50$ steps).

\mypar{BEND without prompts}
An unconditional model has no prompt and thus no retain images that belong to a target. We take in their place the $k = 10$ retain images nearest to the forget image, i.e., the images over which ReTrack computes its weights, and redirect to their mean. The potential is the one of Section~4 of the main paper, applied to the prediction error in pixel space. At $32 \times 32$ pixels the second band contains no frequency, which leaves six band gates and the exponent. The band energies of the normalization (Appendix~\ref{app:potential}) are measured on the first batch of each run.

\mypar{The search}
The search is the one of the main paper and again starts from the squared-$L^2$ loss, $\psi = 0$. One evaluation of a candidate potential is one unlearning run per forget image, followed by the evaluation of the benchmark on fewer generations. The meta-objective asks for forgetting down to the level that ReTrack reaches and measures quality relative to ReTrack:
\begin{multline}
\mathcal{L}_{\mathrm{meta}} = \frac{\mathrm{FID}}{\mathrm{FID}^{\mathrm{R}}} + \frac{(\mathrm{IS}^{\mathrm{R}} - \mathrm{IS})_+}{\mathrm{IS}^{\mathrm{R}}} \\
+ \mu \Big[ (\mathrm{SSCD} - \mathrm{SSCD}^{\mathrm{R}})_+ + \frac{(\mathrm{NLL}^{\mathrm{R}} - \mathrm{NLL})_+}{\mathrm{NLL}^{\mathrm{R}}} + f \Big].
\label{eq:meta-ddpm}
\end{multline}
Here the superscript $\mathrm{R}$ marks the value of ReTrack for the same forget images, seeds, and number of generations, $f$ is the frequency of the T-shirt in percent, $\mu = 4$ as in the main paper, and a term is dropped where the benchmark does not have the metric. All quantities are means over the runs of one evaluation, i.e., over the forget images of a half or over the seeds. ReTrack itself scores $1$. As before, forgetting beyond the level asked for is not rewarded.

\addcontentsline{toc}{subsection}{MNIST T-Shirt}
\mypar{MNIST T-Shirt}
MNIST has a single forget target, so there are no halves to split. Instead, the potential is searched for on ten training seeds that are disjoint from the ten seeds we report, with $10{,}000$ generations per run. The search finds nothing here. The squared-$L^2$ loss with the mean target scores $1.22$ on the fitting seeds, the best candidate of the first step ($g_0 - 1.5$) scores $1.005$, and moving along the descent direction makes the score worse. No candidate therefore reaches the $1$ of ReTrack, and we keep the best one.

Table~\ref{tab:mnist} shows the result on the ten reported seeds, which is a tie. Both methods remove the T-shirt from all $50{,}000$ generations in every seed, and they agree in NLL. In FID and IS the paired differences, $+1.22$ $[-0.10, +3.37]$ and $-0.11$ $[-0.35, +0.04]$, include zero. The mean FID of BEND is carried by one seed ($14.7$), for which ReTrack also has its highest value ($4.5$); the medians are $2.22$ and $2.10$. A second candidate with almost the same score on the fitting seeds, $g_3 + 1.5$ with $1.016$, gives an FID of $2.35 \pm 0.95$. To check the tie, we ran both methods with $20$ further seeds ($10{,}000$ generations). Over all $30$ seeds the paired differences are $+0.50$ $[-0.31, +1.39]$ in FID, $+0.03$ $[0.00, +0.07]$ in NLL, and $-0.05$ $[-0.14, +0.02]$ in IS; the FID exceeds $5$ in one seed of ReTrack and in two of BEND, and no seed of either method generates the T-shirt. In other words, once the forget image is gone from all generations, what this benchmark measures is the variation of a fine-tuning run of $50$ steps, and it does not separate the two methods.

The lower part of Table~\ref{tab:mnist} repeats the two ablations of \citet{S-shi2026retrack} for BEND. The number $k$ of neighbors behind the mean has no effect that stands clear of the variation between seeds: the paired differences in FID to $k = 10$ are $+0.19$ $[-0.82, +1.41]$, $-0.32$ $[-2.18, +1.05]$, and $-0.90$ $[-3.00, +0.57]$ for $k = 1$, $100$, and $1000$. This agrees with what they report for ReTrack. Without the retain term the model is destroyed (FID $127$).

\begin{table}[t]
\centering
\small
\setlength{\tabcolsep}{2.2pt}
\begin{tabular}{lcccc}
\toprule
 & Freq. & & & \\
Method & (\%) $\downarrow$ & NLL $\uparrow$ & FID $\downarrow$ & IS $\uparrow$ \\
\midrule
ReTrack & $0.0000$ & $7.75 \pm 0.05$ & $2.45 \pm 0.93$ & $9.49 \pm 0.11$ \\
BEND & $0.0000$ & $7.77 \pm 0.06$ & $3.67 \pm 3.74$ & $9.38 \pm 0.31$ \\
\midrule
Squared-$L^2$ & $0.0000$ & $7.81 \pm 0.05$ & $2.52 \pm 0.97$ & $9.49 \pm 0.14$ \\
$g_3 + 1.5$ & $0.0000$ & $7.64 \pm 0.05$ & $2.35 \pm 0.95$ & $9.46 \pm 0.13$ \\
$k = 1$ & $0.0000$ & $7.70 \pm 0.07$ & $3.86 \pm 3.37$ & $9.40 \pm 0.27$ \\
$k = 100$ & $0.0000$ & $7.83 \pm 0.08$ & $3.35 \pm 1.69$ & $9.43 \pm 0.15$ \\
$k = 1000$ & $0.0000$ & $7.85 \pm 0.07$ & $2.77 \pm 1.20$ & $9.51 \pm 0.12$ \\
No retain term & $0.0000$ & $8.83 \pm 0.05$ & $127.2 \pm 4.1$ & $6.60 \pm 0.10$ \\
\bottomrule
\end{tabular}
\caption{MNIST T-Shirt (mean $\pm$ standard deviation over ten seeds; $50{,}000$ generations). Lower part: BEND with one component changed, namely another loss (the squared-$L^2$ loss, and the second candidate of the search), another number $k$ of neighbors behind the target, and no retain term ($\lambda = 1$).}
\label{tab:mnist}
\end{table}

\mypar{CIFAR-10: what there is to forget}
The forgetting measure of this benchmark is the SSCD similarity between a forget image and its reconstruction by the model after noising the image to a quarter of the schedule. We measured the same quantity with the pretrained model for images it has never seen. For $500$ images of the CIFAR-10 test set it is $0.52$, for $500$ other training images $0.55$, and for the ten forget images $0.54$ (standard deviation over the images $0.09$). The forget images are therefore reconstructed barely better than images that were never in the training set, i.e., by this measure there is little to remove. ReTrack lowers the similarity to $0.38$, far below the level of unseen images, and lowering it further adds little that this measure can show. We keep the measure, since it is the one of the benchmark, but it limits what the benchmark can show.

\mypar{CIFAR-10: the search}
The ten forget images are split into two halves of five, and every image is unlearned with the potential found on the other half. We first ran the search with a cheaper evaluation, namely $5000$ generations sampled with $100$ steps instead of $1000$, and this search was misled. It closed the coarse bands and put the weight of the loss on the fine ones, with a meta-objective of $0.88$ and $0.87$ on the two halves. Under the protocol of the benchmark these potentials forget more than ReTrack but generate worse images (last row of Table~\ref{tab:cifar}). With $100$ steps every model generates poor images (FID about $70$), and the differences between candidates in this regime do not carry over to $1000$ steps. We report this as a caution: the evaluation inside the search has to use the sampler of the benchmark, as it does on the other datasets.

We then scored the candidates again with the sampler of the benchmark ($2000$ generations with $1000$ steps per run): the squared-$L^2$ loss, the best single move of the first search, and four steps along its descent direction (Table~\ref{tab:cifar-select}). No candidate reaches the $1$ of ReTrack on either half, and the meta-objective rises with the step along the direction: the potentials that put more weight on the fine bands forget more, which is not rewarded, and lose quality. The candidates kept are the single move $g_4 + 1.5$ on half A and the smallest step along the direction on half B. Both stay close to the squared-$L^2$ loss, i.e., on this benchmark the search finds only losses close to it.

\addcontentsline{toc}{subsection}{CIFAR-10: the result}
\mypar{CIFAR-10: the result}
Table~\ref{tab:cifar} gives the result under the protocol of the benchmark, with every image unlearned under the potential of the other half. BEND forgets more than ReTrack by both measures of the benchmark: the paired differences are $+0.11$ $[+0.08, +0.15]$ in NLL, higher for all ten images, and $-0.054$ $[-0.092, -0.022]$ in SSCD, lower for nine. Quality is the same within the variation between images, with paired differences of $+0.35$ $[-0.20, +0.94]$ in FID and $+0.05$ $[-0.01, +0.12]$ in IS. The additional forgetting comes from the potential and not from the target: with the squared-$L^2$ loss the mean target forgets as much as ReTrack ($-0.009$ $[-0.028, +0.009]$ in SSCD). As discussed above, ReTrack already lowers the similarity below that of images the model has never seen, so this measure has little room to reward the additional forgetting.

\begin{table}[t]
\centering
\small
\setlength{\tabcolsep}{4pt}
\begin{tabular}{lcc}
\toprule
Candidate & Half A & Half B \\
\midrule
ReTrack & $1.000$ & $1.000$ \\
Mean target, squared-$L^2$ (start) & $1.013$ & $1.034$ \\
\midrule
Single move $g_4 + 1.5$ & $1.010$ & $1.010$ \\
Along the direction, largest move $0.5$ & $1.013$ & $1.003$ \\
\quad largest move $1$ & $1.020$ & $1.042$ \\
\quad largest move $1.5$ & $1.037$ & $1.088$ \\
\quad largest move $3$ & $1.099$ & $1.217$ \\
\bottomrule
\end{tabular}
\caption{The selection on CIFAR-10 with the sampler of the benchmark: meta-objective~\eqref{eq:meta-ddpm} on the five images of each half (lower is better; ReTrack is $1$ by definition). Kept: $g_4 + 1.5$ on half A and the largest move $0.5$ on half B.}
\label{tab:cifar-select}
\end{table}

\begin{table}[t]
\centering
\footnotesize
\setlength{\tabcolsep}{1.2pt}
\begin{tabular}{lcccc}
\toprule
Method & NLL $\uparrow$ & SSCD $\downarrow$ & FID $\downarrow$ & IS $\uparrow$ \\
\midrule
ReTrack & $3.48 \pm 0.34$ & $0.377 \pm 0.063$ & $8.50 \pm 0.39$ & $9.46 \pm 0.15$ \\
BEND & $3.60 \pm 0.36$ & $0.323 \pm 0.067$ & $8.84 \pm 1.01$ & $9.51 \pm 0.13$ \\
\midrule
Squared-$L^2$ & $3.48 \pm 0.36$ & $0.368 \pm 0.054$ & $8.65 \pm 0.52$ & $9.43 \pm 0.16$ \\
Low-pass & $3.44 \pm 0.37$ & $0.397 \pm 0.057$ & $8.73 \pm 0.83$ & $9.56 \pm 0.19$ \\
$100$-step search & $3.75 \pm 0.36$ & $0.265 \pm 0.065$ & $13.23 \pm 3.55$ & $9.37 \pm 0.16$ \\
\bottomrule
\end{tabular}
\caption{CIFAR-10 (mean $\pm$ standard deviation over the ten forget images; $10{,}000$ generations with $1000$ sampling steps). Lower part: the mean target with other losses, namely the squared-$L^2$ loss, the hand-set low-pass, and the potentials of the search that was evaluated with $100$ sampling steps. \citet{S-shi2026retrack} report $3.44$, $0.37$, $8.45$, and $9.42$ for ReTrack.}
\label{tab:cifar}
\end{table}

\mypar{CelebA-HQ: the search}
As on CIFAR-10, the ten identities are split into two halves of five, and every identity is unlearned with the potential found on the other half. The candidates are scored with $1024$ generations per run and the sampler of the benchmark. On both halves the lowest meta-objective is reached by the same single move of the first step, the gate of the fifth band raised by $1.5$ ($g_4 + 1.5$), with $0.935$ and $0.964$ against the $1$ of ReTrack, where the mean target with the squared-$L^2$ loss scores $1.289$ and $0.981$; the steps along the descent direction are worse than this single move. The potentials of the two halves are therefore the same.

\addcontentsline{toc}{subsection}{CelebA-HQ: the result}
\mypar{CelebA-HQ: the result}
Table~\ref{tab:celeba} gives the result under the protocol of the benchmark. BEND forgets more than ReTrack by both measures: the paired differences are $+0.57$ $[+0.31, +0.87]$ in NLL, higher for all ten identities, and $-0.097$ $[-0.184, -0.023]$ in SSCD, lower for eight. At the same time FID is lower, by $1.7$ $[0.1, 3.3]$, for seven of the ten identities. As on CIFAR-10, the additional forgetting comes from the potential and not from the target. With the squared-$L^2$ loss the mean target forgets as much as ReTrack (paired differences $-0.025$ $[-0.102, +0.040]$ in SSCD and $-0.019$ $[-0.104, +0.045]$ in NLL), and its FID is lower by $1.2$ $[0.0, 2.6]$. Note that the similarity after unlearning scatters widely over the identities (between $0.14$ and $0.70$), so a single run per identity determines it only roughly.

\begin{table}[t]
\centering
\small
\setlength{\tabcolsep}{3pt}
\begin{tabular}{lccc}
\toprule
Method & NLL $\uparrow$ & SSCD $\downarrow$ & FID $\downarrow$ \\
\midrule
ReTrack & $1.52 \pm 0.27$ & $0.393 \pm 0.170$ & $25.4 \pm 2.5$ \\
BEND & $2.09 \pm 0.69$ & $0.296 \pm 0.099$ & $23.8 \pm 2.1$ \\
Mean, squared-$L^2$ & $1.50 \pm 0.22$ & $0.368 \pm 0.152$ & $24.2 \pm 1.8$ \\
\bottomrule
\end{tabular}
\caption{CelebA-HQ (mean $\pm$ standard deviation over the ten forget identities; $50$ sampling steps; $10{,}000$ generations, as in the protocol). \citet{S-shi2026retrack} report $1.51$, $0.41$, and $21.26$ for ReTrack under the protocol.}
\label{tab:celeba}
\end{table}

{
    \small
    \IfFileExists{supp.bbl}{\input{supp.bbl}}{}
}
\clearpage
\makeatletter\immediate\closeout\arxiv@suppaux\let\@auxout\@mainaux\makeatother
\end{document}

%% file: preamble.tex
\usepackage{cuted}    %
\usepackage{caption}  %

\usepackage{algorithm}
\usepackage{algpseudocode}

\input{math}

\newcommand{\mypar}[1]{\paragraph{#1.}}

%% file: math.tex
\newcommand{\va}{{\mathbf{a}}}
\newcommand{\vb}{{\mathbf{b}}}

\newcommand{\vu}{{\mathbf{u}}}
\newcommand{\vv}{{\mathbf{v}}}
\newcommand{\vx}{{\mathbf{x}}}
\newcommand{\vy}{{\mathbf{y}}}

\newcommand{\vzero}{{\mathbf{0}}}

\newcommand{\vepsilon}{{\boldsymbol{\epsilon}}}
\newcommand{\vtheta}{{\boldsymbol{\theta}}}

\newcommand{\mI}{{\mathbf{I}}}

\newcommand{\sA}{{\mathcal{A}}}
\newcommand{\sS}{{\mathcal{S}}}

\providecommand{\R}{\mathbb{R}}
\providecommand{\E}{\mathbb{E}}

\newcommand{\Normal}{\mathcal{N}}

%% file: main.bbl
\begin{thebibliography}{40}
\providecommand{\natexlab}[1]{#1}
\providecommand{\url}[1]{\texttt{#1}}
\expandafter\ifx\csname urlstyle\endcsname\relax
  \providecommand{\doi}[1]{doi: #1}\else
  \providecommand{\doi}{doi: \begingroup \urlstyle{rm}\Url}\fi

\bibitem[Alberti et~al.(2025)Alberti, Hasanaliyev, Shah, and
  Ermon]{alberti2025siss}
Silas Alberti, Kenan Hasanaliyev, Manav Shah, and Stefano Ermon.
\newblock Data unlearning in diffusion models.
\newblock In \emph{Proceedings of the 13th International Conference on Learning
  Representations (ICLR)}, 2025.
\newblock arXiv:2503.01034.

\bibitem[Banerjee et~al.(2005)Banerjee, Guo, and Wang]{banerjee2005bregman}
Arindam Banerjee, Xin Guo, and Hui Wang.
\newblock On the optimality of conditional expectation as a bregman predictor.
\newblock \emph{IEEE Transactions on Information Theory}, 51\penalty0
  (7):\penalty0 2664--2669, 2005.

\bibitem[Bechtle et~al.(2021)Bechtle, Molchanov, Chebotar, Grefenstette,
  Righetti, Sukhatme, and Meier]{bechtle2021learnedloss}
Sarah Bechtle, Artem Molchanov, Yevgen Chebotar, Edward Grefenstette, Ludovic
  Righetti, Gaurav Sukhatme, and Franziska Meier.
\newblock Meta-learning via learned loss.
\newblock In \emph{International Conference on Pattern Recognition (ICPR)},
  2021.

\bibitem[Bregman(1967)]{bregman1967relaxation}
L.~M. Bregman.
\newblock The relaxation method of finding the common point of convex sets and
  its application to the solution of problems in convex programming.
\newblock \emph{USSR Computational Mathematics and Mathematical Physics},
  7\penalty0 (3):\penalty0 200--217, 1967.

\bibitem[Bui et~al.(2025)Bui, Vu, Vuong, Le, Montague, Abraham, Kim, and
  Phung]{bui2025fantastic}
Anh Bui, Thuy-Trang Vu, Long Vuong, Trung Le, Paul Montague, Tamas Abraham,
  Junae Kim, and Dinh Phung.
\newblock Fantastic targets for concept erasure in diffusion models and where
  to find them.
\newblock In \emph{International Conference on Learning Representations
  (ICLR)}, 2025.

\bibitem[Carlini et~al.(2023)Carlini, Hayes, Nasr, Jagielski, Sehwag, Tram\`er,
  Balle, Ippolito, and Wallace]{carlini2023extracting}
Nicholas Carlini, Jamie Hayes, Milad Nasr, Matthew Jagielski, Vikash Sehwag,
  Florian Tram\`er, Borja Balle, Daphne Ippolito, and Eric Wallace.
\newblock Extracting training data from diffusion models.
\newblock In \emph{32nd USENIX Security Symposium (USENIX Security)}, 2023.

\bibitem[Chen et~al.(2024)Chen, Liu, and Xu]{chen2024memorizationfree}
Chen Chen, Daochang Liu, and Chang Xu.
\newblock Towards memorization-free diffusion models.
\newblock In \emph{Proceedings of the IEEE/CVF Conference on Computer Vision
  and Pattern Recognition (CVPR)}, 2024.

\bibitem[Choi et~al.(2022)Choi, Lee, Shin, Kim, Kim, and Yoon]{choi2022p2}
Jooyoung Choi, Jungbeom Lee, Chaehun Shin, Sungwon Kim, Hyunwoo Kim, and
  Sungroh Yoon.
\newblock Perception prioritized training of diffusion models.
\newblock In \emph{Proceedings of the IEEE/CVF Conference on Computer Vision
  and Pattern Recognition (CVPR)}, 2022.

\bibitem[{Commission Nationale de l'Informatique et des
  Libert\'es}(2022)]{cnil2022clearview}
{Commission Nationale de l'Informatique et des Libert\'es}.
\newblock Facial recognition: 20 million euros penalty against {Clearview AI},
  2022.
\newblock Press release of 20 October 2022 on deliberation SAN-2022-019.

\bibitem[Efron and Tibshirani(1993)]{efron1993bootstrap}
Bradley Efron and Robert~J. Tibshirani.
\newblock \emph{An Introduction to the Bootstrap}.
\newblock Chapman \& Hall, New York, 1993.

\bibitem[Fan et~al.(2024)Fan, Liu, Zhang, Wong, Wei, and Liu]{fan2024salun}
Chongyu Fan, Jiancheng Liu, Yihua Zhang, Eric Wong, Dennis Wei, and Sijia Liu.
\newblock {SalUn}: Empowering machine unlearning via gradient-based weight
  saliency in both image classification and generation.
\newblock In \emph{International Conference on Learning Representations
  (ICLR)}, 2024.

\bibitem[Gandikota et~al.(2023)Gandikota, Materzy\'nska, Fiotto-Kaufman, and
  Bau]{gandikota2023esd}
Rohit Gandikota, Joanna Materzy\'nska, Jaden Fiotto-Kaufman, and David Bau.
\newblock Erasing concepts from diffusion models.
\newblock In \emph{Proceedings of the IEEE/CVF International Conference on
  Computer Vision (ICCV)}, 2023.

\bibitem[Gao et~al.(2025)Gao, Lu, Zhou, Chu, Zhang, Jia, Zhang, Fan, and
  Zhang]{gao2025eraseanything}
Daiheng Gao, Shilin Lu, Wenbo Zhou, Jiaming Chu, Jie Zhang, Mengxi Jia, Bang
  Zhang, Zhaoxin Fan, and Weiming Zhang.
\newblock Eraseanything: Enabling concept erasure in rectified flow
  transformers.
\newblock In \emph{International Conference on Machine Learning (ICML)}, 2025.

\bibitem[George et~al.(2026)George, Murata, Takida, Mopuri, and
  Mitsufuji]{george2026tilde}
Naveen George, Naoki Murata, Yuhta Takida, Konda~Reddy Mopuri, and Yuki
  Mitsufuji.
\newblock {TILDE}: {TIL}t-based distributional erasure for concept unlearning,
  2026.
\newblock arXiv:2607.06432.

\bibitem[Golatkar et~al.(2020)Golatkar, Achille, and
  Soatto]{golatkar2020neggrad}
Aditya Golatkar, Alessandro Achille, and Stefano Soatto.
\newblock Eternal sunshine of the spotless net: Selective forgetting in deep
  networks.
\newblock In \emph{Proceedings of the IEEE/CVF Conference on Computer Vision
  and Pattern Recognition (CVPR)}, 2020.

\bibitem[Hang et~al.(2023)Hang, Gu, Li, Bao, Chen, Hu, Geng, and
  Guo]{hang2023minsnr}
Tiankai Hang, Shuyang Gu, Chen Li, Jianmin Bao, Dong Chen, Han Hu, Xin Geng,
  and Baining Guo.
\newblock Efficient diffusion training via {Min-SNR} weighting strategy.
\newblock In \emph{Proceedings of the IEEE/CVF International Conference on
  Computer Vision (ICCV)}, 2023.

\bibitem[Heng and Soh(2023)]{heng2023selective}
Alvin Heng and Harold Soh.
\newblock Selective amnesia: A continual learning approach to forgetting in
  deep generative models.
\newblock In \emph{Advances in Neural Information Processing Systems 36
  (NeurIPS)}, 2023.

\bibitem[Heusel et~al.(2017)Heusel, Ramsauer, Unterthiner, Nessler, and
  Hochreiter]{heusel2017fid}
Martin Heusel, Hubert Ramsauer, Thomas Unterthiner, Bernhard Nessler, and Sepp
  Hochreiter.
\newblock {GANs} trained by a two time-scale update rule converge to a local
  {Nash} equilibrium.
\newblock In \emph{Advances in Neural Information Processing Systems 30
  (NeurIPS)}, 2017.

\bibitem[Ho et~al.(2020)Ho, Jain, and Abbeel]{ho2020ddpm}
Jonathan Ho, Ajay Jain, and Pieter Abbeel.
\newblock Denoising diffusion probabilistic models.
\newblock In \emph{Advances in Neural Information Processing Systems 33
  (NeurIPS)}, 2020.

\bibitem[Jiang et~al.(2021)Jiang, Dai, Wu, and Loy]{jiang2021focal}
Liming Jiang, Bo Dai, Wayne Wu, and Chen~Change Loy.
\newblock Focal frequency loss for image reconstruction and synthesis.
\newblock In \emph{Proceedings of the IEEE/CVF International Conference on
  Computer Vision (ICCV)}, 2021.

\bibitem[Karras et~al.(2022)Karras, Aittala, Aila, and Laine]{karras2022edm}
Tero Karras, Miika Aittala, Timo Aila, and Samuli Laine.
\newblock Elucidating the design space of diffusion-based generative models.
\newblock In \emph{Advances in Neural Information Processing Systems
  (NeurIPS)}, 2022.

\bibitem[Khalafi et~al.(2026)Khalafi, Ribeiro, and Ding]{khalafi2026erasure}
Shervin Khalafi, Alejandro Ribeiro, and Dongsheng Ding.
\newblock Unlearning in diffusion models: A unified framework with {KL}
  divergence and likelihood constraints.
\newblock In \emph{International Conference on Machine Learning (ICML)}, 2026.

\bibitem[Kingma and Gao(2023)]{kingma2023understanding}
Diederik~P. Kingma and Ruiqi Gao.
\newblock Understanding diffusion objectives as the {ELBO} with simple data
  augmentation.
\newblock In \emph{Advances in Neural Information Processing Systems
  (NeurIPS)}, 2023.

\bibitem[Kumari et~al.(2023)Kumari, Zhang, Wang, Shechtman, Zhang, and
  Zhu]{kumari2023ablating}
Nupur Kumari, Bingliang Zhang, Sheng-Yu Wang, Eli Shechtman, Richard Zhang, and
  Jun-Yan Zhu.
\newblock Ablating concepts in text-to-image diffusion models.
\newblock In \emph{IEEE/CVF International Conference on Computer Vision
  (ICCV)}, 2023.

\bibitem[Lipman et~al.(2023)Lipman, Chen, Ben-Hamu, Nickel, and
  Le]{lipman2023flowmatching}
Yaron Lipman, Ricky T.~Q. Chen, Heli Ben-Hamu, Maximilian Nickel, and Matt Le.
\newblock Flow matching for generative modeling.
\newblock In \emph{International Conference on Learning Representations
  (ICLR)}, 2023.

\bibitem[Lipman et~al.(2024)Lipman, Havasi, Holderrieth, Shaul, Le, Karrer,
  Chen, Lopez-Paz, Ben-Hamu, and Gat]{lipman2024fmguide}
Yaron Lipman, Marton Havasi, Peter Holderrieth, Neta Shaul, Matt Le, Brian
  Karrer, Ricky T.~Q. Chen, David Lopez-Paz, Heli Ben-Hamu, and Itai Gat.
\newblock Flow matching guide and code, 2024.

\bibitem[Liu et~al.(2022)Liu, Ye, Wright, Stone, and Liu]{liu2022bome}
Bo Liu, Mao Ye, Stephen Wright, Peter Stone, and Qiang Liu.
\newblock Bome! bilevel optimization made easy: A simple first-order approach.
\newblock In \emph{Advances in Neural Information Processing Systems 35
  (NeurIPS)}, 2022.

\bibitem[Panda et~al.(2026)Panda, {Varun M.~S.}, Jain, Maharana, and
  {Prathosh}]{panda2024vdu}
Subhodip Panda, {Varun M.~S.}, Shreyans Jain, Sarthak~Kumar Maharana, and A.~P.
  {Prathosh}.
\newblock Unlearning in diffusion models under data constraints: A variational
  inference approach.
\newblock \emph{Transactions on Machine Learning Research (TMLR)}, 2026.

\bibitem[Park and Park(2026)]{park2025frequency}
Jinseong Park and Mijung Park.
\newblock Not every time and frequency need to be forgotten in diffusion
  unlearning.
\newblock In \emph{ICML Workshop on Foundations of Deep Generative Models:
  Understanding Memorization, Generalization, and Reasoning}, 2026.
\newblock arXiv:2510.17917.

\bibitem[Pizzi et~al.(2022)Pizzi, Roy, Ravindra, Goyal, and
  Douze]{pizzi2022sscd}
Ed Pizzi, Sreya~Dutta Roy, Sugosh~Nagavara Ravindra, Priya Goyal, and Matthijs
  Douze.
\newblock A self-supervised descriptor for image copy detection.
\newblock In \emph{Proceedings of the IEEE/CVF Conference on Computer Vision
  and Pattern Recognition (CVPR)}, 2022.

\bibitem[Ren et~al.(2024)Ren, Li, Zeng, Xu, Lyu, Xing, and
  Tang]{ren2024unveiling}
Jie Ren, Yaxin Li, Shenglai Zeng, Han Xu, Lingjuan Lyu, Yue Xing, and Jiliang
  Tang.
\newblock Unveiling and mitigating memorization in text-to-image diffusion
  models through cross attention.
\newblock In \emph{European Conference on Computer Vision (ECCV)}, 2024.

\bibitem[Rombach et~al.(2022)Rombach, Blattmann, Lorenz, Esser, and
  Ommer]{rombach2022ldm}
Robin Rombach, Andreas Blattmann, Dominik Lorenz, Patrick Esser, and Bj\"orn
  Ommer.
\newblock High-resolution image synthesis with latent diffusion models.
\newblock In \emph{Proceedings of the IEEE/CVF Conference on Computer Vision
  and Pattern Recognition (CVPR)}, 2022.

\bibitem[Salimans et~al.(2016)Salimans, Goodfellow, Zaremba, Cheung, Radford,
  and Chen]{salimans2016improved}
Tim Salimans, Ian Goodfellow, Wojciech Zaremba, Vicki Cheung, Alec Radford, and
  Xi Chen.
\newblock Improved techniques for training {GANs}.
\newblock In \emph{Advances in Neural Information Processing Systems
  (NeurIPS)}, 2016.

\bibitem[Shi et~al.(2026)Shi, Jin, Zhang, and Gu]{shi2026retrack}
Qitan Shi, Cheng Jin, Jiawei Zhang, and Yuantao Gu.
\newblock {ReTrack}: Data unlearning in diffusion models through redirecting
  the denoising trajectory.
\newblock In \emph{Proceedings of the 29th International Conference on
  Artificial Intelligence and Statistics (AISTATS)}, 2026.
\newblock arXiv:2509.13007.

\bibitem[Somepalli et~al.(2023{\natexlab{a}})Somepalli, Singla, Goldblum,
  Geiping, and Goldstein]{somepalli2023diffusion}
Gowthami Somepalli, Vasu Singla, Micah Goldblum, Jonas Geiping, and Tom
  Goldstein.
\newblock Diffusion art or digital forgery? {Investigating} data replication in
  diffusion models.
\newblock In \emph{Proceedings of the IEEE/CVF Conference on Computer Vision
  and Pattern Recognition (CVPR)}, 2023{\natexlab{a}}.

\bibitem[Somepalli et~al.(2023{\natexlab{b}})Somepalli, Singla, Goldblum,
  Geiping, and Goldstein]{somepalli2023understanding}
Gowthami Somepalli, Vasu Singla, Micah Goldblum, Jonas Geiping, and Tom
  Goldstein.
\newblock Understanding and mitigating copying in diffusion models.
\newblock In \emph{Advances in Neural Information Processing Systems
  (NeurIPS)}, 2023{\natexlab{b}}.

\bibitem[{U.S. Federal Trade Commission}(2021)]{ftc2021everalbum}
{U.S. Federal Trade Commission}.
\newblock In the matter of {Everalbum, Inc.}: Decision and order, 2021.
\newblock Docket No.\ C-4743, issued May 6, 2021.

\bibitem[Wang et~al.(2023)Wang, Chan, and Loy]{wang2023clipiqa}
Jianyi Wang, Kelvin C.~K. Chan, and Chen~Change Loy.
\newblock Exploring {CLIP} for assessing the look and feel of images.
\newblock In \emph{Proceedings of the 37th AAAI Conference on Artificial
  Intelligence}, 2023.

\bibitem[Wen et~al.(2024)Wen, Liu, Chen, and Lyu]{wen2024detecting}
Yuxin Wen, Yuchen Liu, Chen Chen, and Lingjuan Lyu.
\newblock Detecting, explaining, and mitigating memorization in diffusion
  models.
\newblock In \emph{International Conference on Learning Representations
  (ICLR)}, 2024.

\bibitem[Wu et~al.(2025)Wu, Le, Hayat, and Harandi]{wu2024erasediff}
Jing Wu, Trung Le, Munawar Hayat, and Mehrtash Harandi.
\newblock Erasing undesirable influence in diffusion models.
\newblock In \emph{Proceedings of the IEEE/CVF Conference on Computer Vision
  and Pattern Recognition (CVPR)}, 2025.

\end{thebibliography}


\begin{thebibliography}{15}
\providecommand{\natexlab}[1]{#1}
\providecommand{\url}[1]{\texttt{#1}}
\expandafter\ifx\csname urlstyle\endcsname\relax
  \providecommand{\doi}[1]{doi: #1}\else
  \providecommand{\doi}{doi: \begingroup \urlstyle{rm}\Url}\fi

\bibitem[Banerjee et~al.(2005)Banerjee, Guo, and Wang]{S-banerjee2005bregman}
Arindam Banerjee, Xin Guo, and Hui Wang.
\newblock On the optimality of conditional expectation as a bregman predictor.
\newblock \emph{IEEE Transactions on Information Theory}, 51\penalty0
  (7):\penalty0 2664--2669, 2005.

\bibitem[Bi{\'n}kowski et~al.(2018)Bi{\'n}kowski, Sutherland, Arbel, and
  Gretton]{S-binkowski2018demystifying}
Miko{\l}aj Bi{\'n}kowski, Danica~J. Sutherland, Michael Arbel, and Arthur
  Gretton.
\newblock Demystifying {MMD} {GANs}.
\newblock In \emph{International Conference on Learning Representations
  (ICLR)}, 2018.

\bibitem[Chen et~al.(2024)Chen, Mo, Hou, Wu, Liao, Sun, Yan, and
  Lin]{S-chen2024topiq}
Chaofeng Chen, Jiadi Mo, Jingwen Hou, Haoning Wu, Liang Liao, Wenxiu Sun, Qiong
  Yan, and Weisi Lin.
\newblock {TOPIQ}: A top-down approach from semantics to distortions for image
  quality assessment.
\newblock \emph{IEEE Transactions on Image Processing}, 2024.

\bibitem[Efron and Tibshirani(1993)]{S-efron1993bootstrap}
Bradley Efron and Robert~J. Tibshirani.
\newblock \emph{An Introduction to the Bootstrap}.
\newblock Chapman \& Hall, New York, 1993.

\bibitem[Heusel et~al.(2017)Heusel, Ramsauer, Unterthiner, Nessler, and
  Hochreiter]{S-heusel2017fid}
Martin Heusel, Hubert Ramsauer, Thomas Unterthiner, Bernhard Nessler, and Sepp
  Hochreiter.
\newblock {GANs} trained by a two time-scale update rule converge to a local
  {Nash} equilibrium.
\newblock In \emph{Advances in Neural Information Processing Systems 30
  (NeurIPS)}, 2017.

\bibitem[Ho et~al.(2020)Ho, Jain, and Abbeel]{S-ho2020ddpm}
Jonathan Ho, Ajay Jain, and Pieter Abbeel.
\newblock Denoising diffusion probabilistic models.
\newblock In \emph{Advances in Neural Information Processing Systems 33
  (NeurIPS)}, 2020.

\bibitem[Karpathy and Fei-Fei(2015)]{S-karpathy2015deep}
Andrej Karpathy and Li Fei-Fei.
\newblock Deep visual-semantic alignments for generating image descriptions.
\newblock In \emph{Proceedings of the IEEE Conference on Computer Vision and
  Pattern Recognition (CVPR)}, 2015.

\bibitem[Ke et~al.(2021)Ke, Wang, Wang, Milanfar, and Yang]{S-ke2021musiq}
Junjie Ke, Qifei Wang, Yilin Wang, Peyman Milanfar, and Feng Yang.
\newblock {MUSIQ}: Multi-scale image quality transformer.
\newblock In \emph{Proceedings of the IEEE/CVF International Conference on
  Computer Vision (ICCV)}, 2021.

\bibitem[Lin et~al.(2014)Lin, Maire, Belongie, Hays, Perona, Ramanan,
  Doll{\'a}r, and Zitnick]{S-lin2014coco}
Tsung-Yi Lin, Michael Maire, Serge Belongie, James Hays, Pietro Perona, Deva
  Ramanan, Piotr Doll{\'a}r, and C.~Lawrence Zitnick.
\newblock Microsoft {COCO}: Common objects in context.
\newblock In \emph{European Conference on Computer Vision (ECCV)}, 2014.

\bibitem[Lipman et~al.(2024)Lipman, Havasi, Holderrieth, Shaul, Le, Karrer,
  Chen, Lopez-Paz, Ben-Hamu, and Gat]{S-lipman2024fmguide}
Yaron Lipman, Marton Havasi, Peter Holderrieth, Neta Shaul, Matt Le, Brian
  Karrer, Ricky T.~Q. Chen, David Lopez-Paz, Heli Ben-Hamu, and Itai Gat.
\newblock Flow matching guide and code, 2024.

\bibitem[Oquab et~al.(2024)Oquab, Darcet, Moutakanni, Vo, Szafraniec, Khalidov,
  Fernandez, Haziza, Massa, El-Nouby, et~al.]{S-oquab2024dinov2}
Maxime Oquab, Timoth{\'e}e Darcet, Th{\'e}o Moutakanni, Huy~V. Vo, Marc
  Szafraniec, Vasil Khalidov, Pierre Fernandez, Daniel Haziza, Francisco Massa,
  Alaaeldin El-Nouby, et~al.
\newblock {DINOv2}: Learning robust visual features without supervision.
\newblock \emph{Transactions on Machine Learning Research (TMLR)}, 2024.

\bibitem[Park and Park(2026)]{S-park2025frequency}
Jinseong Park and Mijung Park.
\newblock Not every time and frequency need to be forgotten in diffusion
  unlearning.
\newblock In \emph{ICML Workshop on Foundations of Deep Generative Models:
  Understanding Memorization, Generalization, and Reasoning}, 2026.
\newblock arXiv:2510.17917.

\bibitem[Salimans et~al.(2016)Salimans, Goodfellow, Zaremba, Cheung, Radford,
  and Chen]{S-salimans2016improved}
Tim Salimans, Ian Goodfellow, Wojciech Zaremba, Vicki Cheung, Alec Radford, and
  Xi Chen.
\newblock Improved techniques for training {GANs}.
\newblock In \emph{Advances in Neural Information Processing Systems
  (NeurIPS)}, 2016.

\bibitem[Shi et~al.(2026)Shi, Jin, Zhang, and Gu]{S-shi2026retrack}
Qitan Shi, Cheng Jin, Jiawei Zhang, and Yuantao Gu.
\newblock {ReTrack}: Data unlearning in diffusion models through redirecting
  the denoising trajectory.
\newblock In \emph{Proceedings of the 29th International Conference on
  Artificial Intelligence and Statistics (AISTATS)}, 2026.
\newblock arXiv:2509.13007.

\bibitem[Zhang et~al.(2018)Zhang, Isola, Efros, Shechtman, and
  Wang]{S-zhang2018lpips}
Richard Zhang, Phillip Isola, Alexei~A. Efros, Eli Shechtman, and Oliver Wang.
\newblock The unreasonable effectiveness of deep features as a perceptual
  metric.
\newblock In \emph{Proceedings of the IEEE/CVF Conference on Computer Vision
  and Pattern Recognition (CVPR)}, 2018.

\end{thebibliography}
